\documentclass[
]{ceurart}

\usepackage{moreverb,url}
\usepackage{setspace}
\usepackage{caption} 
\usepackage{graphicx}
\usepackage{float}
\usepackage{url}
\usepackage{rotating}
\usepackage{enumitem}
\usepackage{multirow}
\usepackage{longtable}
\usepackage{pdflscape}
\usepackage{comment}
\usepackage[table,xcdraw]{xcolor}
\usepackage{listings}
\usepackage{booktabs} 
\usepackage{footmisc}

\usepackage{listings}
\begin{document}

\title{ONTrust: A Reference Ontology of Trust}

\author[1]{Glenda Amaral}[%
email=g.c.mouraamaral@utwente.nl,
]
\cormark[1]

\address[1]{Semantics, Cybersecurity \& Services, University of Twente, The Netherlands}
\address[2]{Industrial Engineering and Business Information Systems, University of Twente, The Netherlands}
\address[3]{DAFIST, University of Genoa, Italy}

\author[1]{Tiago Prince Sales}[%
email=t.princesales@utwente.nl,
]

\author[3]{Riccardo Baratella}[%
email=baratellariccardo@gmail.com,
]

\author[3]{Daniele Porello}[%
email=daniele.porello@unige.it,
]

\author[2]{Renata Guizzardi}[%
email=r.guizzardi@utwente.nl,
]

\author[1]{Giancarlo Guizzardi}[%
email=g.guizzardi@utwente.nl,
]

\cortext[1]{Corresponding author.}

\begin{abstract}
Trust has stood out more than ever in the light of recent innovations. Some examples are advances in artificial intelligence that make machines more and more humanlike, and the introduction of decentralized technologies (e.g. blockchains), which creates new forms of (decentralized) trust. These new developments have the potential to improve the provision of products and services, as well as to contribute to individual and collective well-being. However, their adoption depends largely on trust. In order to build trustworthy systems, along with defining laws, regulations and proper governance models for new forms of trust, it is necessary to properly conceptualize trust, so that it can be understood both by humans and machines. This paper is the culmination of a long-term research program of providing a solid ontological foundation on trust, by creating reference conceptual models to support information modeling, automated reasoning, information integration and semantic interoperability tasks.   
To address this, a Reference Ontology of Trust (ONTrust) was developed, grounded on the Unified Foundational Ontology and specified in OntoUML, which has been applied in several initiatives, to demonstrate, for example, how it can be used for conceptual modeling and enterprise architecture design, for language evaluation and (re)design, for trust management, for requirements engineering, and for trustworthy artificial intelligence (AI) in the context of affective Human-AI teaming. ONTrust formally characterizes the concept of trust and its different types, describes the different factors that can influence trust, as well as explains how risk emerges from trust relations. To illustrate the working of ONTrust, the ontology is applied to model two case studies extracted from the literature. 
\end{abstract}

\begin{keywords}
  Trust \sep
  Unified Foundational Ontology \sep
  OntoUML
\end{keywords}

\maketitle

\section{Introduction}
Trust is at the at the heart of most everyday interactions. Still, it is a rather puzzling phenomenon: How to build trust? How to earn it back? How to assess whether someone is trustworthy? How can someone show that is trustworthy? Despite the undoubtable importance of trust, there are no pre-defined rules or formulas to answer these questions. 

The term trust has been used to refer to different types of relationships, such as the trust between individuals, as well as between individuals and organizations, individuals and autonomous agents, between software systems operating in a network, trust in the context of offline or online commercial relationships, among others. In recent years, the importance of trust has become even more evident due to new technological advances, such as the ones we have seen in the field of artificial intelligence (AI) \cite{lukyanenko2022trust,yang2022user,gille2020we}. Innovations in AI have made it possible to build increasingly advanced software systems, such as systems capable of diagnosing patients, deciding whether to grant a loan, driving cars, and so on. However, their adoption depends largely on trust. People will only be able to fully benefit from these technologies if they can trust them \cite{hleg2019ethics,glikson2020human,hidalgo2021humans,regona2026building}. Another example is the introduction of blockchains and other decentralized technologies \cite{zetzsche2020decentralized,schar2021decentralized}, which create a new form of (decentralized) trust that requires the definition of laws, regulations, and the creation of proper governance models \cite{ur2019trust}. 

In all of the aforementioned cases, the concept of trust is fundamental. Therefore, in order to build trustworthy systems, be able to define laws and proper governance models to decentralized trust, properly assess if someone (or something) is trustworthy, one first needs to understand what trust means. Although much progress has been made to clarify the ontological nature of trust, the term remains overloaded and there is not yet a shared or prevailing, and conceptually clear definition for it \cite{castelfranchi2010trust, mcknight2001trust}. In the light of the above, this paper advocates for the need of a reference ontology of trust to serve as a basis for communication, consensus and alignment among different approaches and perspectives, as well as to foster interoperability across heterogeneous application domains.

This paper is part of a long-term research program of providing a solid conceptual foundation on trust and related concepts \cite{amaral2019towards,amaral2021ROT,amaral2022trustCBDC, baratella2023many}
to support information modeling, automated reasoning, information integration and semantic interoperability tasks. In this direction, a Reference Ontology of Trust (ONTrust\footnote{The term ROT was previously used to refer to the Reference Ontology of Trust \cite{amaral2019towards}, henceforth termed ONTrust.}) was proposed in a previous effort. ONTrust is a reference model grounded on the Unified Foundational Ontology \cite{guizzardi2005ontological, guizzardi2021ufo} and specified in OntoUML\footnote{OntoUML is a conceptual modeling language whose primitives reflect the ontological distinctions put forth by the UFO ontology.}, that formally characterizes the concept of trust and its different types, describes the different factors that can influence trust, as well as explains how risk emerges from trust relations. Besides the theoretical work, ONTrust has been applied in several initiatives, in different areas, to demonstrate, for example, how it can be used for conceptual modeling and enterprise architecture design \cite{amaral2020TPL}, for language evaluation and (re)design \cite{amaral2020TPL}, for trust management \cite{amaral2022trustCBDC}, for requirements engineering \cite{amaral2020trustworthiness, amaral2021trustworthinessPix}, for the development of more effective security awareness models \cite{oliveira2025ontological}, for decision making \cite{amaral2025unpacking}, for trustworthy artificial intelligence (AI) in the context of affective Human-AI teaming \cite{amaral2025combining}, among others \cite{touameur2025guitares,ding2024conceptual,shishkov2023incorporating,lanasri2020trust}.

This paper extends the above-mentioned previous work by: (i) validating the ontology via visual model simulation, using the Alloy Analyzer \cite{benevides2010validating}
; (ii) defining an axiomatization to increase the ontology’s precision; and (iii) revisiting the ontology to discuss some special cases, such as \textit{self-trust}, \textit{trust in types} and to provide a deeper account of the concept of institution-based trust. In addition, ONTrust is applied to model two case studies from the literature, both to validate the ontology expressivity and its capacity to represent real-world situations, and to demonstrate its usefulness and relevance.

As previously mentioned, the ontology representation language chosen to model ONTrust is OntoUML \cite{guizzardi2005ontological}, a version of UML class diagrams that has been designed such that its modeling primitives reflect the ontological distinctions put forth by UFO (see section \ref{sec:ufo}), and its grammatical constraints follow UFO axiomatization. Our motivation for choosing OntoUML is twofold. Firstly, because of its expressivity. OntoUML is an ontologically well-founded conceptual modeling language, supported by expressive logical theories, which allows for the representation of the subject domain with truthfulness, clarity and expressivity (regardless of computational requirements). Secondly because the \textit{OntoUML Toolkit} contains a rich set of tools to facilitate the ontology engineering process, such as ontological design patterns and anti-patterns \cite{guizzardi2011design}, visual model simulation \cite{benevides2010validating}, and transformations for codification technologies  \cite{barcelos2013automated, rybola2016towards,gufo2020}, which drive the implementation of lightweight ontologies supported by efficient computational algorithms (e.g., OWL-DL, RDF, Alloy), thus satisfying the need for computationally-aware implementations.

This paper is organized as follows. Section \ref{sec:ufo} introduces the reader to the main notions of UFO. Then, Section \ref{sec:on-trust} discusses the ontological nature of trust, considering the different interpretations found in the literature. Section \ref{sec:rot-ontology} presents the Reference Ontology of Trust. In Section \ref{sec:case-studies}, ONTrust is applied to model two case studies from the literature: (i) the case of trust in IT-mediated elections in Brazil \cite{avgerou2013explaining}; and (ii) a case of trust in artificial intelligence for medical diagnosis \cite{juravle2020trust}. Section \ref{sec:related-work} presents some related work and Section \ref{sec:conclusions} concludes the paper.

\section{Ontological Foundations}
\label{sec:ufo}

This paper provides an ontological analysis of \textit{trust}, grounded on the Unified Foundational Ontology (UFO). UFO is an axiomatic domain independent formal theory, developed based on a number of theories from Formal Ontology, Philosophical Logics, Philosophy of Language, Linguistics and Cognitive Psychology. UFO is divided into three incrementally layered compliance sets: UFO-A, an ontology of endurants (objects) \cite{guizzardi2005ontological}, UFO-B, an ontology of events (perdurants) \cite{guizzardi2013towards}, and UFO-C, an ontology of social entities built on the top of UFO-A and UFO-B, which addresses terms related to the spheres of intentional and social things \cite{guizzardi2010ontology,guizzardi2008grounding,guizzardi2021ufo}. For an in-depth discussion and formalization, one should refer to \cite{guizzardi2005ontological,guizzardi2013towards}. UFO is the theoretical basis of OntoUML, a language for Ontology-driven Conceptual Modeling that has been successfully employed in a number of academic and industrial projects in several domains, such as services, value, petroleum and gas, media asset management, telecommunications, and government \cite{guizzardi2015towards}. OntoUML was designed such that its modeling primitives reflect the ontological distinctions of its underlying ontology, and its grammar is enriched with semantically-motivated syntactical constrains that mirror UFO’s axiomatization. UFO is also formally connected to a number set of tools to facilitate the ontology engineering process, such as ontological design patterns and anti-patterns \cite{guizzardi2011design}, visual model simulation \cite{benevides2010validating}, automated model diagnosis and repair via learning \cite{fumagalli2020towards}, and transformations for codification technologies \cite{barcelos2013automated, rybola2016towards}. In particular, UFO has a partial translation to OWL termed gUFO \cite{gufo2020}, which is suitable for knowledge graph applications. The motivation for using UFO is to provide an accessible and sharable modelling of trust that may be applied across domains to foster the interoperability and the mutual understanding among modellers.

UFO makes a fundamental distinction between \textsf{individuals} (particulars), and \textsf{types} (or universals), that is, patterns of features that are repeatable across individuals. Individuals can be \textsf{concrete} or \textsf{abstract}. Concrete individuals are further categorized into \textsf{endurants} (roughly, things or object-like entities) or \textsf{perdurants} (roughly, events, occurrences, processes). Within the category of endurants, UFO distinguishes \textsf{substantials} and \textsf{moments} (also termed aspects, abstract particulars, or variable tropes \cite{moltmann202019}). Substantials are existentially independent objects, such as the Moon, an enterprise, a person, a horse. Moments are existentially dependent entities and can only exist by inhering in other entities such as Alice's capacity to swim (which depends on her). Moments are distinguished into \textsf{intrinsic moments} and \textsf{relators}.

Intrinsic moments are existentially dependent on a single individual. These include \textsf{qualities}, i.e., reifications of categorical properties such as height, weight, age, electrical charge, color, and \textsf{modes}. Modes can bear their own moments, including their own qualities, which can vary in independent ways (e.g. John's headache, Alice’s capacity to swim). The category of modes includes dispositions (e.g., functions, capabilities, vulnerabilities) as well as externally dependent entities (e.g., the trust of citizens in politicians, the commitment of John towards Mary to watch a film on Netflix on Saturday night). 

\textsf{Externally dependent modes} inhere in an entity while being externally dependent on another entity. For example, the love of a mother for her child is a mode that inheres in the mother and externally depends on the child. Other examples are beliefs, desires, intentions, perceptions, symptoms, among many others.

\textsf{Relators} are moments existentially dependent on multiple individuals. They are individuals with the power of connecting entities. For example, an enrollment relator connects an individual playing the Student role with an Educational Institution. Every instance of a relator type is existentially dependent on at least two distinct entities. Moreover, relators are typically composed of modes, for example, in the way that the service agreement between John and Netflix is composed of their mutual commitments and claims. 

\begin{figure*}
\setlength{\fboxsep}{0pt}%
\setlength{\fboxrule}{0pt}%
\begin{center}
    \centering
    \includegraphics[width=\textwidth]{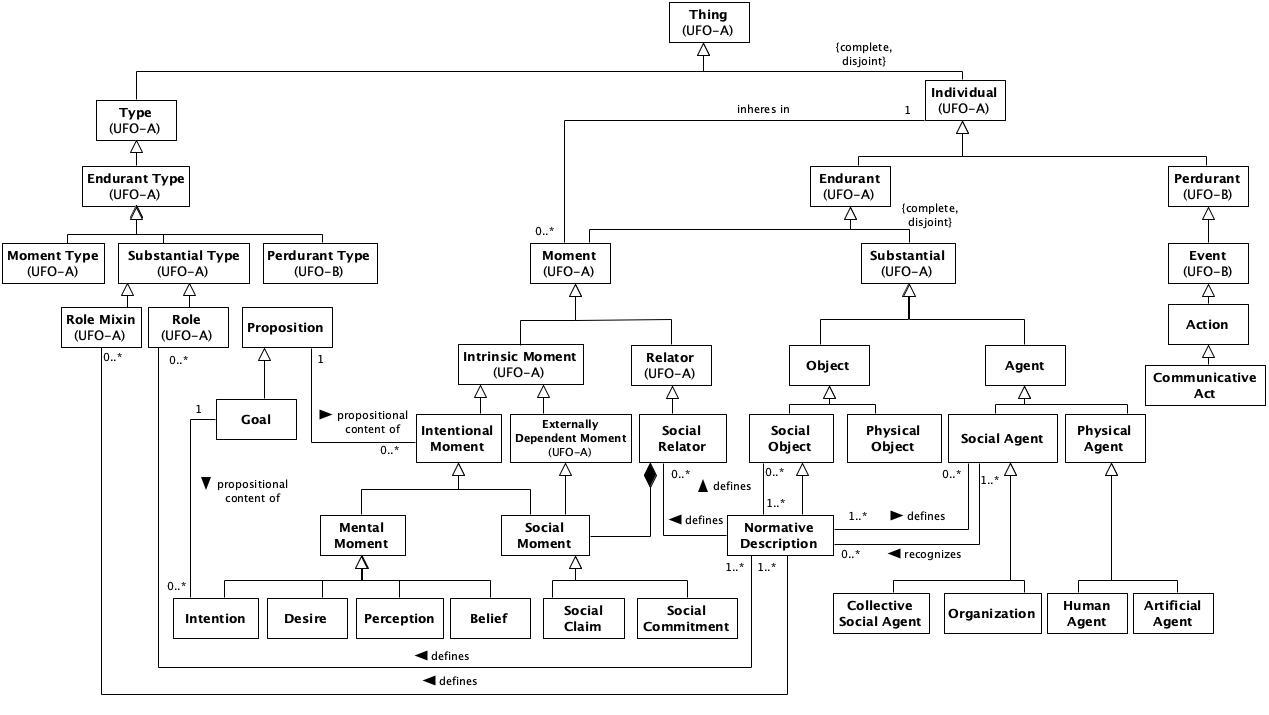}
    \caption{Fragment of UFO-C.}
    \label{fig:ufo-c}
    \end{center}
\end{figure*}

In our analysis, we shall rely mainly on some concepts defined in UFO-C \cite{guizzardi2008grounding,guizzardi2010ontology} (Figure \ref{fig:ufo-c}). A basic distinction in UFO-C is between \textsf{agents} and (non-agentive) \textsf{objects}. An agent is a substantial that creates actions and to which we can ascribe mental states (intentional moments). Agents can be \textsf{physical} (e.g., a person) or \textsf{social} (e.g., an organization). A distinction is made between \textsf{human agent} and \textsf{artificial agent} (both subkinds of physical agent), to differentiate human agents from software (or hardware) agents. An object, on the other hand, is a substantial unable to perceive events or to have intentional moments. Objects can also be further categorized into \textsf{physical} (e.g., a book, a car) and \textsf{social objects} (e.g., money, language). 

Agents can bear special kinds of moments named \textsf{intentional moments}. Intentional moments can be \textsf{social moments} or \textsf{mental moments}. Mental moments refer to the capacity of some properties of certain individuals to refer to actual or potential situations of reality \cite{guizzardi2008grounding}. A \textsf{Mental Moment} is existentially dependent on a particular \textsf{Agent}, being an inseparable part of its mental state (Figure \ref{fig:ufo-c}). Examples of mental moments include \textsf{perceptions}, \textsf{beliefs}, \textsf{desires} and \textsf{intentions}. Perceptions express how agents sense their environment and the things that happen around them. Beliefs have a propositional content that agents consider to be true. They can be justified by situations in reality. Examples include my belief that Rome is the Capital of Italy, and the belief that the Moon orbits the Earth. Beliefs can be formed by perceptions expressing how agents sense their environment and the things that happen around them. Desires and intentions can be fulfilled or frustrated. A desire expresses the will of an agent towards a possible situation (e.g., a desire that Brazil wins the next World Cup). Intentions (or internal commitments) express desired states of affairs for which the agent commits to pursuing directly or indirectly, by pursuing the coming about of a certain state of affairs (e.g., Mary’s intention of paying a bill using an internet banking system; Bob's intention not to pay the bill at the restaurant by having the intention of running away from it). The propositional content of an intention is a \textsf{goal}. Intentions cause agents to perform \textsf{Actions}, which are intentional events, i.e., events that have the specific purpose of satisfying some intention. In particular, a \textsf{Communicative Act} (a speech act such as inform, ask or promise) \cite{searle1995construction} is an example of action.

Communicative Acts can be used to create \textsf{Social Moments}. In this view, language
not only represents reality but also creates a part of reality \cite{searle1995construction}. Thus, social moments are types of intentional moments that are created by the exchange of communicative acts and the consequences of these exchanges (e.g., goal adoption, delegation). For instance, suppose that Mary subscribes to Netflix. When signing a business agreement, she performs a communicative act (a promise). This act creates a \textsf{Social Commitment} towards that organization: a commitment to pay for the subscribed plan (the propositional content). Moreover, it also creates a Social Claim of that organization towards Mary with respect to that particular propositional content. Briefly speaking, social commitment is a commitment of an agent A towards another agent B. As an externally dependent moment, a social commitment inheres in A and is externally dependent on B. The social commitments necessarily cause the creation of an internal commitment in A. Also, associated to this internal commitment, a social claim of B towards A is created. \textsf{Commitments} and \textsf{claims} always form a pair that are about a unique propositional content. For example, when John rents a car at a car rental office, he commits (a social commitment towards that organization) to pay the rental fee according to the table of ``vehicle categories and prices''. Thus, it creates a social claim of the rental car office towards John with respect to this particular propositional content (the payment). Finally, a social relator (e.g., a ``service agreement'') mediates a relation between two or more individuals (e.g., ``Mary'', and ``Netflix'') that play different roles in the relation (e.g., ``service customer'', and ``service provider''). By participating in a social relator, the individuals bear a number of social commitments and claims.

A first observation about commitments is that, as previously mentioned, only agents can make commitments and satisfy claims. Ordinary objects are passive entities with no such capabilities. Therefore, when it comes to objects, we may adopt a kind of indirect commitment, which refer to the commitment of the producer or responsible for this object (a human agent or an organization). 

A further observation is regarding artificial agents, which can be seen as powerful abstractions for capturing human’s beliefs and intentions, to model some of their reasoning capabilities and interactions, and for capturing the commitments they establish with others on a human agent or an organization's behalf. In general, artificial agents implement services provided by a person or an organization, which are regulated by contracts established by the commitments between agents. This leads to the need of mechanisms to ensure responsibility and accountability \cite{hleg2019ethics} for artificial agents and their outcomes, both before and after their development, deployment and use.

Finally, a normative description is a type of social object that defines one or more rules/norms recognized by at least one social agent and that can define social intrinsic and relational properties (e.g., social commitment types), social objects (e.g., the crown of the King of Spain) and social roles (e.g., president, or pedestrian). Examples of normative descriptions include the Italian Constitution and the University of Twente PhD program regulations. 

\section{On Trust}
\label{sec:on-trust}

A wide number of definitions of trust have been proposed along the years, across several areas, such as psychology \cite{rotter1967new,tyler2006people,mujdricza2019roots,eyal2024trust}, sociology \cite{barber1983logic,gambetta2000can,luhmann2018trust}, philosophy \cite{simpson2023trust,simon2020routledge}, economics \cite{williamson1993calculativeness}, law \cite{cross2004law}, and more recently, computer science \cite{giorgini2005modeling,moyano2012conceptual,afroogh2023probabilistic}. Although much progress has been made to clarify the nature of trust, the term remains semantically overloaded and there is not yet a shared or prevailing, and conceptually clear notion of trust \cite{pytlikzillig2016consensus,castelfranchi2010trust}. 

A classic definition of trust, widely accepted in the literature, was proposed by the sociologist Diego Gambetta, who defines trust as ``the subjective probability with which an agent expects that another agent or group of agents will perform a particular action on which its welfare depends'' \cite{gambetta2000can}. In his definition it is clear the existence of both a \textit{trustor} and a \textit{trustee}, as well as a belief of the trustor about the behavior of the trustee. Gambetta also relates trust to an intention of the trustor regarding her welfare and the uncertainty about the trustee's behavior, which reveals the existence of a certain degree of risk. In fact, this idea that trust presupposes a situation of risk is ubiquitous in the literature. For instance, Luhmann \cite{luhmann2018trust} argues that when people trust  others, they act ``as if they knew the future'', and uncertainty is transformed into risk. Also, Castelfranchi and Falcone \cite{castelfranchi2010trust} state that without uncertainty and risk there is no trust.

A similar concept of trust is proposed by Mayer et al. \cite{mayer1995integrative}, who define trust as ``the willingness of a party to be vulnerable to the actions of another party, based on the expectation that the other party will perform a particular action important to the trustor, irrespective of the ability to monitor or control that other party''. Also here, the authors refer to the expectations (or beliefs) of the trustor regarding the trustee and correlates trust to the trustor's goals (the actions of the other party that are important to the trustor). According to the authors, by trusting another party, the trustor makes herself vulnerable and exposed to the occurrence of risk events.

Rousseau and colleagues relied on a large interdisciplinary literature and on the identification of fundamental and convergent elements to define trust as ``a psychological state of a trustor comprising the intention to accept vulnerability in a situation involving risk, based on positive expectations of the intentions or behavior of the trustee'' \cite{rousseau1998not}. Note that, also in this definition, the authors reinforce the presence of the trustor's expectations regarding the trustee, as well as the relationship between trust and risk: by trusting, the trustor accepts to become vulnerable to the trustee in terms of potential failure of the expected action and result, as the trustee may not perform the expected action or the action may not have the desired result. 

McKnight and Chervany \cite{mcknight2001trust} compared sixty-five definitions of trust from different sources to propose an interdisciplinary model of conceptual trust types that takes into account several important aspects of trust and some of their mutual interactions. For example, the authors are able to distinguish between a belief and a behavioral component of trust, and to explain that the latter depends on the former. The belief component is related to cognitive perceptions about the attributes or characteristics of others, i.e., the trustor believes, with ``feelings of relative security'', that the trustee is willing and able to act in her interest. The behavioral component means that a person voluntarily takes actions that makes herself dependent on another entity, with a feeling of relative security, even though negative consequences are possible. According to the authors, trust-related behavior comes in a number of subconstruct forms because many actions can make one dependent on another, such as cooperation, information sharing, informal agreements, decreasing controls, accepting influence, granting autonomy, and transacting business. 

A further important aspect in the model of McKnight and Chervany \cite{mcknight2001trust} is the distinction between \textit{interpersonal trust} and \textit{institution-based trust}. This distinction is also made by Luhmann \cite{luhmann2018trust}, who defines:

\begin{itemize}
    \item \textbf{interpersonal trust} as that between individuals that frequently have face-to-face contact and become familiar with each other without substantially taking recourse to institutional arrangements; and
    
    \item \textbf{institution-based trust} as that in the reliable functioning of certain social systems, which no longer refers to a reality the trustor is acquainted with, but is built on impersonal and generalized ``media of communication'', such as the monetary system and the legal system. 
\end{itemize}
 
According to McKnight and Chervany \cite{mcknight2001trust}, institution-based trust affects interpersonal trust by ``making the trustor feel more comfortable about trusting others, as she securely believes that protective structures (such as guarantees, contracts, regulations, promises, legal recourse, processes or procedures) are in place that are conducive to situational success''. For example, people believe in the efficacy of a bank to take care of their money because of the existence of laws and institutions that insure them against loss. Lewis and Weigert  \cite{lewis1985trust} argue that institution-based trust is indispensable for the effective functioning of ``symbolic media of exchange'', such as money and political power. They argue that ``without public trust in the reliability, effectiveness, and legitimacy of money, laws, and other cultural symbols, modern social institutions would soon disintegrate''.

More recently, Castelfranchi and Falcone  \cite{castelfranchi2010trust} analyzed the concept of trust as a composed and ``layered'' notion, relying on some key aspects: (i) a mental attitude and a disposition towards another agent; (ii) a decision and intention to rely upon the other, which makes the trustor vulnerable; (iii) the act of relying upon the trustee's expected behavior; and (iv) the consequent overt social interaction and relation between the trustor and the trustee.

In their definition of trust, Castelfranchi and Falcone \cite{castelfranchi2010trust} emphasize the role of the trustor's goal by stating that an ``agent trusts another only relative to a goal, i.e., for something she wants to achieve, that she desires or needs''. They also reinforce the idea of trust consisting of beliefs about the trustee and her behavior: ``the belief that the trustee is able and willing to do the needed action; the belief that the trustee will appropriately do the action, as the trustor wishes; and the belief that the trustor can make herself less defended and more vulnerable''. As for the behavioral component of trust, Castelfranchi and Falcone \cite{castelfranchi2010trust} argue that there may be mental trust without the corresponding behavioral part (i.e., without an action). That may happen because the level of trust is not sufficient; the level of trust is sufficient, but there are other reasons preventing the action (e.g. prohibitions); or trust is just potential, i.e., a predisposition (e.g. ``the trustor would, might rely on the trustee, if/when…, but it is not (yet) the case''). 

The distinction between the mental attitude from the behavioral aspect of trust also appears in Kelsall's non‑reliance commitment account of trust \cite{kelsall2024towards}, which proposes two different concepts, namely trust and trusting reliance. According to this distinction, to trust is to believe that a trustee is fit for trusting reliance with respect to some object or objects of trust (a mental attitude). As for trust reliance, Kelsall defines that X trusting relies Y only if X believes Y to have a commitment to Z and X relies upon Y to keep that commitment. According to this author, trusting reliance is the special form of reliance that conforms to the trusting attitude.

In summary, what can be extracted from these different proposals is that there is a conceptual core to be enlightened in order to properly define trust. Therefore, to conceptualize trust, one must refer to: (i) \textit{agents and their goals}; (ii) \textit{agents' beliefs}; (iii) possibly executable \textit{actions} of a given type; and (iv) \textit{risk}.

\section{The Reference Ontology of Trust}
\label{sec:rot-ontology}

This section presents the Reference Ontology of Trust (ONTrust)
\footnote{The complete version of ONTrust in OntoUML and its implementation in OWL are available at \url{http://purl.org/krdb-core/trust-ontology}.} \cite{amaral2019towards,amaral2021ROT,amaral2022trustCBDC,baratella2023many},
 a well-founded ontology that formalizes the concept of trust as discussed in the previous section. The mental aspects of trust are formalized in the OntoUML
\footnote{Once more, OntoUML is a conceptual modeling language whose primitives reflect the ontological distinctions put forth by the UFO ontology \cite{guizzardi2005ontological}} 
 models depicted in Figures \ref{fig:ground-trust}, \ref{fig:social-weak-strong-trust} and \ref{fig:institution-based-trust}, and the particular behavioral aspect of trust, as well as the relation between trust and risk in Figure \ref{fig:trust-risk}. In the OntoUML diagrams depicting ONTrust, types are represented in purple, objects in pink, events in yellow, moments in blue, relators in green and situations in orange.

\subsection{The Many Flavors of Trust}

The term \textit{trust} is commonly used as an umbrella for different kinds of relations. In order to provide a clear and sound account for the notion of trust, we distinguish and characterize different levels of trust. First, we identify a core set of characteristics that are peculiar to what we call \textit{Ground Trust}, presented in section \ref{sec:ground-trust}. Then, in sections \ref{sec:social-trust} to \ref{sec:trusted-delegation} we discuss additional properties and relations that differentiate further other kinds of trust and their connections, namely \textit{Social Trust}, \textit{Weak Trust}, \textit{Strong Trust}, \textit{Institution-based Trust} and \textit{Trusted Delegation}.

\subsubsection{Ground Trust}
\label{sec:ground-trust}

\begin{figure*}
\setlength{\fboxsep}{0pt}%
\setlength{\fboxrule}{0pt}%
\begin{center}
    \centering
    \includegraphics[width=\textwidth]{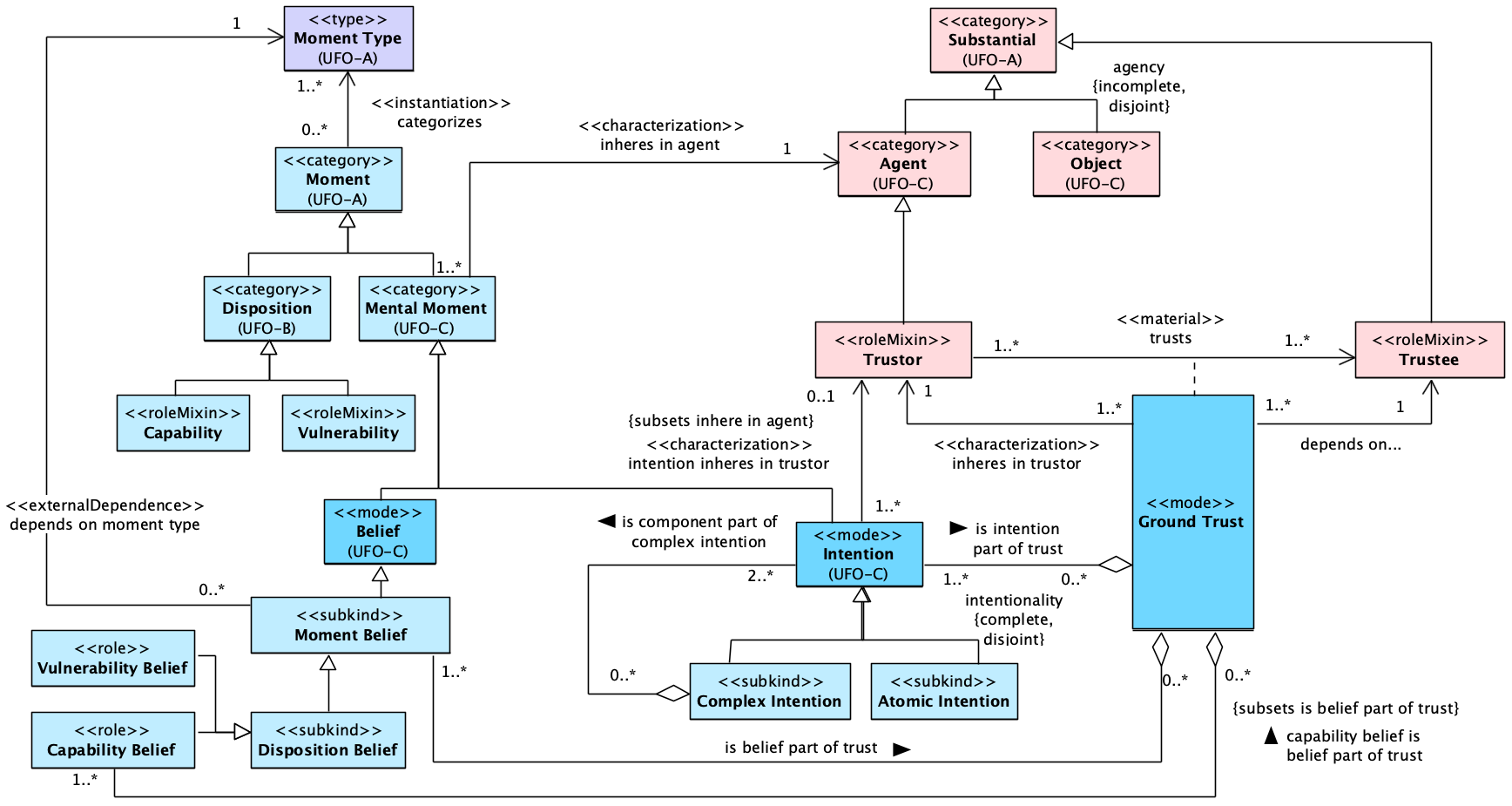}
    \caption{Ground Trust.}
    \label{fig:ground-trust}
    \end{center}
\end{figure*}

In ONTrust, \textsf{Ground Trust} (Figure \ref{fig:ground-trust}) is defined as a complex mental state of a \textsf{Trustor} agent, composed of a set of \textsf{Beliefs} about a \textsf{Trustee} and its behavior. \textsf{Ground Trust} is always about an \textsf{Intention} of the \textsf{Trustor} regarding a goal. Also omissions may be relevant in this context, as an omission can be seen as the (positive) action one performs while omitting \cite{varzi2007omissions}. In this case, the \textsf{Trustor} might precisely rely on the fact that the \textsf{Trustee} will not do a specific action or, more generally, that the \textsf{Trustee} will not do anything at all. 

Note that the \textsf{Intention} of achieving a goal is an essential part of \textsf{Trust}, thus distinguishing it from a mere bunch of beliefs. If the \textsf{Trustor} does not have a goal, she cannot really decide, nor care about something that could depend on the behavior of another entity. For example, stating that the mother believes that the babysitter is capable of taking care of her kids is different from saying that she trusts the babysitter to do so. For the mother's mental state to correspond to trust, the intention of having someone to look after her children must be part of her mental plan.

The \textsf{Trustor} may trust the \textsf{Trustee} regarding a certain \textsf{Intention}, but may not trust it with respect to a different one. For example, I trust my dentist to fix a cavity in my tooth, but not to fix my computer. Furthermore, such an intention is not always atomic, it can instead be a complex intention. While atomic intentions have no proper parts, complex intentions are aggregations of at least two disjoint intentions. For instance, in the trust relation ``Bob trusts a certain airline to take him on his holiday trip comfortably and safely”, trust is about a complex intention, composed of (i) Bob's intention of traveling; (ii) his intention of being safe; and (iii) his intention of being comfortable.  Let us now consider a situation in which Mary trusts an application provider for collecting her location data, except when she is in sensitive places such as a cancer treatment center for her medical treatment, since such  information may lead to disclosing her disease. In this example, trust is also about a complex intention, composed of (i) Mary’s intention of having her location data collected; (ii) Mary's intention of not having her diseased disclosed to others; and (iii) Mary’s intention of preserving her location privacy when she is in sensitive places.  

The \textsf{Trustor} is necessarily an ``intentional entity'', that is, a cognitive agent, an agent endowed with goals and beliefs. In UFO, a belief is a special type of mode, named mental moment, which is existentially dependent on a particular agent, being an inseparable part of its mental state. As for the \textsf{Trustee}, it is an entity capable of impacting one's intentions by the outcome of its behavior \cite{castelfranchi2010trust}, regardless if this involves an action or an omission (e.g. `doing nothing’, `abstaining from doing X’). Moreover, note that the \textsf{Trustee} is not necessarily aware of being trusted. An example, mentioned by Castelfranchi and Falcone \cite{castelfranchi2010trust}, is a person running to catch a bus. Even if this person is not seen by the bus driver and the people waiting for the bus at the stop, she may attribute to these people the intention to take the bus, and thus the intention to stop it. In such a case, the runner is trusting in the people at the bus stop to do so. 

In our ontology, in accordance with Castelfranchi and Falcone's proposal \cite{castelfranchi2010trust}, the \textsf{Trustee} is not necessarily a cognitive system, or an animated or autonomous agent. It can also be a lot of things we rely upon in our daily activity: rules, procedures, conventions, infrastructures, technology and artifacts in general, tools, authorities and institutions, environmental regularities, as well as different types of social systems. When the \textsf{Trustee} is a cognitive agent we have a special type of trust, which we name here \textsf{Social Trust} (see section \ref{sec:social-trust}). 

Another important aspect to consider is the possibility of trusting in types. For example, if I say I trust an adult to lift a heavy object, it means that I trust every individual belonging to the class of adult people to do so. However, this is not necessarily always true, as there may be instances in this class that I do not trust. For example, I might not trust a sick adult to lift a heavy object, because they might be weakened due to their health conditions.  As we understand, trust in types does not necessarily apply, however we consider that beliefs about a particular type can influence trust in some of their instances. For example, I believe that veterinary students are skilled with animals. John is a veterinarian student. Thus, it's very likely that I will trust John to look after my pet lizard when I go on vacation.

It is worth noting that the same individual can play the roles of both \textsf{Trustor} and \textsf{Trustee}, which stands for a case of \textit{self-trust}. For example, I can trust myself to write a paper on trustworthiness requirements for software systems, but it may be the case that I do not trust myself to change a flat tire.

As previoulsly mentioned, trust is a complex mental state that inheres in the \textsf{Trustor} and is dependent on the \textsf{Trustee}. In general, \textsf{Trustor} and \textsf{Trustee} are different individuals and this dependence relation corresponds to a case of external dependence, except for the case of self-trust, in which the external dependence concept does not apply, as \textsf{Trustor} and \textsf{Trustee} are the same individual.

The \textsf{Trust} complex mental state of the \textsf{Trustor} is composed of her \textsf{Intention} and a set of beliefs about the \textsf{Trustee} that we name here \textsf{Moment Beliefs}. We use the notion of \textit{belief} defined in UFO (see section \ref{sec:ufo}) to model \textsf{Moment Belief} as a belief that inheres in the \textsf{Trustor} and is externally dependent on an external entity \cite{guizzardi2013towards,azevedo2015modeling}, more specifically, on types of moments (entity \textsf{Moment Types}) that the \textsf{Trustor} believes inhere in the \textsf{Trustee}, such as capabilities, vulnerabilities and intentions.

As illustrated in Figure \ref{fig:ground-trust}, \textsf{Disposition Belief} is a specialization of \textsf{Moment Belief}, which is a belief about a \textsf{Disposition}\footnotemark{} of the \textsf{Trustee}. It is distinguished into \textsf{Capability Belief} and \textsf{Vulnerability Belief}. The former refers to a belief about a \textsf{Trustee}'s \textsf{Capability}\footnotemark[\value{footnote}] --- the \textsf{Trustor} believes that the \textsf{Trustee} is capable of performing a desired action or exhibiting a desired behavior --- while the latter refers to a \textsf{Trustee}'s \textsf{Vulnerability}\footnotemark[\value{footnote}] \footnotetext{The notions of mode, disposition, capability and vulnerability are defined and discussed in section \ref{sec:ufo}.} --- the \textsf{Trustor} believes that the \textsf{Trustee}’s vulnerabilities will not prevent it from performing the desired action or exhibiting the desired behavior. Note that the very same disposition may play the role of a capability, a vulnerability, or even a threat capability\footnote{Capabilities are usually perceived as beneficial, as they enable the manifestation of events desired by an agent. However, when the manifestation of a capability enables undesired events that threaten an agent’s abilities to achieve a goal, it can be seen as a threat capability \cite{sales2018cover}.}. For example, in the scope of military operations, information can be seen both as a capability (as digital data and networks support and facilitate the achievement of military objectives) and a vulnerability (as confidential information can be disclosed as a consequence of a cyber-attack). For this reason, both capabilities and vulnerabilities are represented as roles of dispositions (Figure \ref{fig:ground-trust}) that inhere in the \textsf{Trustee}, which are manifested in particular situations, through the occurrence of events \cite{guizzardi2013towards}. We adopt here the interpretation of capability proposed by Azevedo \cite{azevedo2015modeling}, who defined capability as the power to bring about a desired outcome. 

Finally, the \textsf{trusts} relation between the \textsf{Trustor} and the \textsf{Trustee} is a relation that is \textit{non-symmetric}, \textit{non-reflexive} and \textit{non-transitive}. An example that evidences the non-symmetry is a child that trusts her father to lift a heavy object, but the father does not trust his child to do so. However, it is possible that the father trusts the mother to take care of their kids and vice-versa. Trust is non-reflexive because an agent may or may not trust herself to perform actions. For example, an athlete may trust herself to run one kilometer in ten minutes, but not to cook a sophisticated meal. Lastly, it is non-transitive because agents might have different evaluations about the same entity's trustworthiness. For instance, it is very well possible that Alice trusts Bob for performing certain actions and Bob trusts Charlie for performing the same actions, but it is not the case the Alice trusts Charlie to perform them. 

\subsubsection{Social Trust}
\label{sec:social-trust}

In ONTrust, \textsf{Social Trust} is a specialization of \textsf{Ground Trust} in which the \textsf{Trustee} is an \textsf{Agent} (Figure \ref{fig:social-weak-strong-trust}). The \textsf{Intention Belief} is specific to this type of trust, as in this case the \textsf{Trustee} is a cognitive agent endowed with goals. \textsf{Intention Belief} is a belief about an \textsf{Intention} of the \textsf{Trustee} --- the \textsf{Trustor} believes that the \textsf{Trustee} intends to perform a desired action or exhibit an expected behavior. For example, a mother who trusts a babysitter to take care of her kids believes that: (i) the babysitter has experience in caring for children and is First Aid trained (a belief about the babysitter’s capabilities); (ii) the babysitter is well and probably is not going to have health issues (a belief about the babysitter’s vulnerabilities); and (iii) the babysitter is willing to take good care of her children (a belief about the babysitter’s intention).

In \textsf{Social Trust} the \textsf{Trustor} believes that the \textsf{Trustee} has the motivation to act toward the goal of the trust relation. In the babysitter example above, the mother believes that the babysitter intends to take good care of her children. It is important to notice that the motivations of the \textsf{Trustee} may be different as they can be – however, to the extent that these motivations allow the \textsf{Trustor} to achieve their goal, they work.

\subsubsection{Weak Trust}
\label{sec:weak-trust}

\textsf{Ground Trust} provides a minimal ground for the \textsf{Trustor}’s belief in the trustworthiness of the \textsf{Trustee}, however it does not include any implicit or explicit commitment of the \textsf{Trustee} toward the \textsf{Trustor}. Nevertheless, it is important to distinguish a relation of Ground Trust, where no commitment is done, from types of trust where either this commitment is done or the \textsf{Trustor} believes that they have done.

\textsf{Weak Trust} is a type of trust relation in which the \textsf{Trustor} has the belief that the \textsf{Trustee} has a commitment to goal c, which the trust relation is about. For instance, Tom trusts his friend Lora to drive him to the airport in time to catch his flight. It is because Tom and Lora are friends that he believes she is committed to accomplishing goal c that is the goal upon which this trust relation is founded. However, this belief may very well be false. It is crucial to note the notion of commitment referred to here is one of social commitment (see section \ref{sec:ufo}), which entails, in the case the belief is true, social claims concerning goal c on part of the \textsf{Trustor}. Further, when the trustee is not a cognitive agent, we shall adopt a kind of indirect commitment, e.g., the commitments of a car or a clock are the commitments of the producer of these items (see section \ref{sec:ufo}).

In ONTrust, \textsf{Weak Trust} is a specialization of \textsf{Social Trust}, in which the \textsf{Trustor} has the belief that the \textsf{Trustee} has a commitment to the goal the trust relation is about (that is, the propositional content of the \textsf{Trustor}’s \textsf{Intention}). 

As illustrated in Figure \ref{fig:social-weak-strong-trust}, \textsf{Social Commitment Belief} is a \textsf{Social Commitment} (see section \ref{sec:ufo}) that represents the \textsf{Trustor}’s belief that the \textsf{Trustee} is committed to the goal, for the achievement of which she counts upon the \textsf{Trustee}. As explained in section \ref{sec:ufo}, \textsf{Social Commitments} are types of \textsf{Moments} (Figure \ref{fig:ufo-c}), and therefore instantiate \textsf{Moment Type}.

\subsubsection{Strong Trust}
\label{sec:strong-trust}

\textsf{Strong Trust} is a specialization of \textsf{Weak trust} in which the \textsf{Trustor} knows that the \textsf{Trustee} explicitly commits herself to goal c, to which the trust relation is about, as part of an existing agreement. For instance, suppose that Tom trusts Anonymous Reviewer to review his article in a fair and careful way. Then, this trust relation presupposes the explicit social commitment of Anonymous Reviewer to review Tom’s article, and Tom knows about this fact. 

Strong Trust also applies in the case of commitments given by artificial agents on behalf of organizations or human agents (see section \ref{sec:ufo}). In this case, accessible mechanisms must be in place to ensure responsibility and accountability for artificial agents’ outcomes. For example, knowing that redress is possible when things go wrong is key to ensure trust \cite{hleg2019ethics}. A recent incident illustrates the issue of trust and commitments made by artificial agents. We refer to the case of Air Canada’s chatbot \cite{Melnick2024}, which promised a discount to a client, but after the purchase, the airline claimed that that the chatbot had been wrong. The customer explained that he made the purchased based on his trust on Air Canada’s chatbot, however the airline argued that the chatbot was a separate legal entity ``responsible for its own actions’’ \cite{Melnick2024}. As we can see, this dispute sheds light on questions about how the issue of accountability should be addressed when commitments are made by artificial agents. In this particular case, the Canadian tribunal came down on the side of the customer, ruling that Air Canada is responsible for all the information on its website, regardless of whether the information comes from a static page or a chatbot.

Another point worth mentioning is that, at least in the case of \textsf{Strong Trust}, it seems that the explicit commitment of the \textsf{Trustee} has consequences also for the commitments, the beliefs, and the acts of the \textsf{Trustor}. For instance, if Mary trusts John to cook dinner and John explicitly commits himself to it, then Mary will act as if she won’t cook the dinner, she is committed not to cook the dinner, and she believes she will not cook the dinner. What is more, if Mary started cooking the dinner, then John may rightly be frustrated, because at least he has the belief that Mary won’t cook the dinner. Thus, in the case of \textsf{Strong Trust}, the commitment of the \textsf{Trustee} seems to generate some commitment on the side of the \textsf{Trustor} too. 

In ONTrust, \textsf{Strong Trust} is modeled as a specialization of \textsf{Weak Trust} that is grounded on an existing \textsf{Agreement} that includes the \textsf{Trustee Commitement} to the goal c to which \textsf{Strong Trust} is about. ONTrust applies the concept of social relator (see section \ref{sec:ufo}) to model \textsf{Agreement} as a relator composed of pairs of social commitments and social claims that inhere either in \textsf{Trustee} or in the \textsf{Trustor}. Briefly speaking, \textsf{Agreement} mediates a relation between the \textsf{Trustor} and \textsf{Trustee} by being a social relator aggregating their explicit commitments and claims. See figure \ref{fig:social-weak-strong-trust} for the modeling of Strong Trust and Agreement.

\subsubsection{Trusted Delegation}
\label{sec:trusted-delegation}

The previous kinds of trust relation did not take into consideration the phenomenon of delegation. Many trust relations involve the fact that the trustor delegates \textit{goal c} of the trust relation to the trustee. 
For instance, suppose that the trustor is the main author of the paper and trusts one of their co-authors to do a part of the paper either because of the lack of time or because the co-author is more versed in this respect. We now turn to the relation between trust and delegation.

Webster dictionary defines delegation as ``the act of empowering to act for another". This is consistent with the concept of delegation adopted here, which is the one argued by Guizzardi R. and Guizzardi G. \cite{guizzardi2010ontology}: a material relation between a delegator and a delegatee, which involves a social relator composed of pairs of commitment/claim and dependence between delegator and delegatee. So defined, delegation clearly does not entail trust. There may be delegation without trust, due to coercion or because one does not have another option but to delegate. For example, suppose one's car breaks down in the middle of nowhere. After a few hours, a driver stops and says he can fix it for 100 dollars, but he has to go grab his tools in another town first and he demands to be paid in advance. In this case, the car owner indeed delegates (thus, creating corresponding commitments and claims) the goal of having the car fixed to the alleged mechanic. The car owner does that because he has no other choice, even if he does not trust the delegatee. 

But what kind of connection is there between Trusted Delegation and different forms of trust? Clearly, Trusted Delegation is compatible with both Weak Trust and Strong Trust. At minimum, if the trustor deliberately delegates to the trustee to achieve \textit{goal c} upon which the trust relation is founded, then the trustor must have the belief that the trustee is committed to achieve \textit{goal c}.
However, it is too strong to require that Trusted Delegation entails Strong Trust. For instance, John trusts his wife to go get their children at school, so he delegates this task to her. However, she might not have explicitly committed herself to this goal. Hence, Trusted Delegation requires at least Weak Trust. In other words, for Trusted Delegation to exist, there must be either Weak Trust or Strong Trust.

\begin{figure*}
\setlength{\fboxsep}{0pt}%
\setlength{\fboxrule}{0pt}%
\begin{center}
    \centering
    \includegraphics[width=\textwidth]{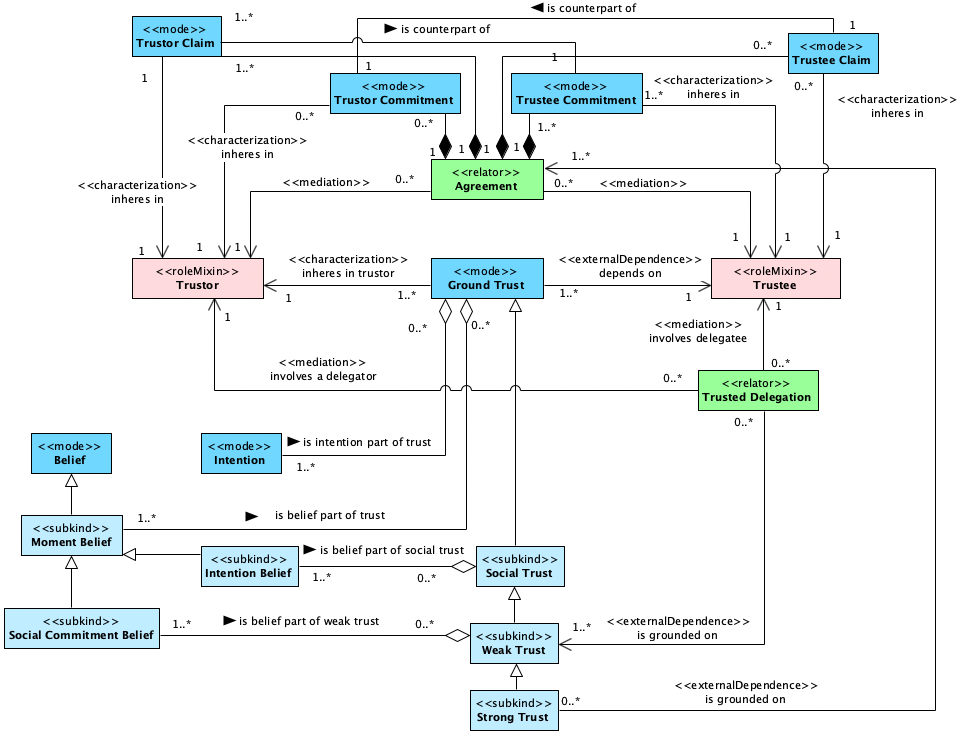}
    \caption{Social Trust, Weak Trust, Strong Trust and Trusted Delegation}
    \label{fig:social-weak-strong-trust}
    \end{center}
\end{figure*}

To account for trust relations in which the trustor delegates the goal of the trust relation to the trustee, ONTrust introduces the concept of \textsf{Trusted Delegation}. We model \textsf{Trusted Delegation} as a social relator, grounded on \textsf{Weak Trust}, which mediates a delegation relation between the \textsf{Trustor} (the delegator) and the \textsf{Trustee} (the delegatee) (see figure \ref{fig:social-weak-strong-trust}).

 \subsubsection{Institution-based Trust}
\label{sec:institution-based-trust}

In the case of \textsf{Institution-based Trust}, the \textsf{Trustee} is  what we name here a \textsf{Social System}. \textsf{Institution-based Trust} builds upon the existence of shared rules, regularities, conventional practices, etc. and is related to a \textsf{Social System Trustee}.  This comes from the sociology tradition positing that people can rely on others because of structures, situations or assigned social roles that provide assurances that things will go well \cite{barber1983logic}. \textsf{Institution-based Trust} refers to beliefs about those protective structures, and not just about the people involved. In this paper we term these protective structures \textsf{Social Systems}. We adopt the interpretation of \textsf{Social Systems} as orderly arrangements of social entities that interact with each other, based on established and prevalent social rules that structure social interactions. \textsf{Social Systems} create a shared world of clear rules and reliable standards, which no longer refers to a personally known reality, but is built on impersonal and generalized ``media of communication'' \cite{luhmann2018trust} such as the monetary system and the legal system. Let us take the example of the monetary system as a \textsf{Social System Trustee}: in society, individuals provide something of value in return for a token they trust to be able to use in the future to obtain something else of value, as well as they trust that the value of the instrument will be stable in terms of goods and services.  An additional example, involving different types of trust, is the case of a person who buys a phone in an e-commerce platform. Here we can identify several trust relations: (i) the buyer's social trust in the seller about her delivering the phone in perfect state; (ii) the buyer's trust in the phone about it behaving as she expects; (iii) the buyer's and the seller's institution-based trust in the monetary system; (iv) the buyer's and the seller's institution-based trust in the legal system (in case one of the parties does not fulfill its commitments); (v) the buyer's and the seller's trust in the online platform.

\begin{figure*}
\setlength{\fboxsep}{0pt}%
\setlength{\fboxrule}{0pt}%
\begin{center}
    \centering
    \includegraphics[width=\textwidth]{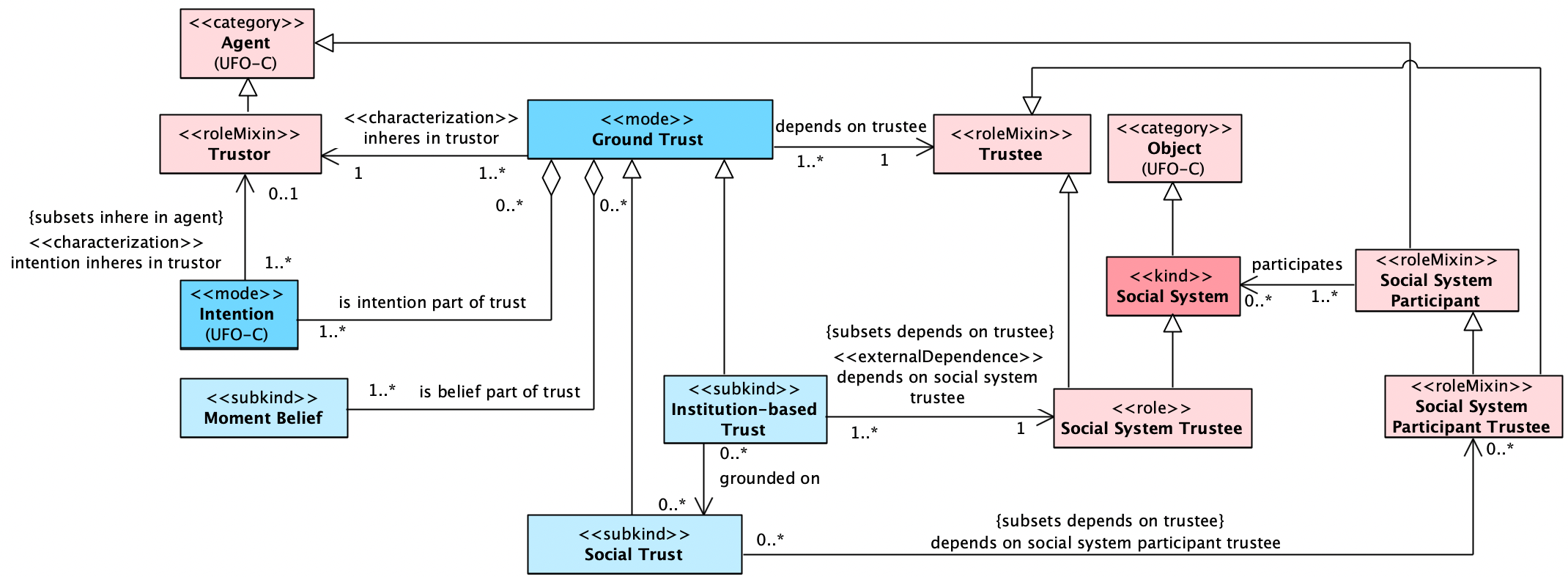}
    \caption{Institution-based Trust}
    \label{fig:institution-based-trust}
    \end{center}
\end{figure*}

In ONTrust, \textsf{Institution-based Trust} is a specialization of \textsf{Ground Trust}, in which the \textsf{Trustee} is a \textit{Social System} (Figure \ref{fig:institution-based-trust}). An important aspect, related to the nature of \textsf{Social Systems}, is that they can be seen as \textit{integral wholes}, whose parts play particular \textit{functional roles} that contribute in specific ways to the functionality of the whole \cite{guizzardi2005ontological,sales2017fleet}. UFO includes micro-theories to address different types of \textit{part-whole relations} \cite{guizzardi2005ontological,sales2017fleet} generally recognized in cognitive science \cite{pribbenow2002meronymic,gerstl1995midwinters}. \textsf{Social Systems} embody one particular kind of such parthood relations, namely, \textit{component-functional complex} \cite{sales2017fleet}. In UFO's terminology, this ``componentOf'' relation is used to relate entities that are \textit{functional complexes}. Some examples of functional complexes are an organization, a legal system or a monetary system and their corresponding ``componentOf'' relations (e.g., presidency-organization, law-legal system, currency-monetary system). Consequently, \textsf{Social Systems} can be defined as functional complexes composed of social entities (e.g. social roles, social objects, social relationships, normative descriptions and so on). An example of \textsf{Social System} is the legal system, which is an integral whole composed of a number of social entities, such as social roles (e.g. lawyer, judge, etc.), social objects (e.g. contract, court sentence), normative descriptions (e.g. laws, regulations) and others that contribute in complementary manners to the functionality of the whole.

It is worth noting that institutional-based trust is also grounded on the trustors' social trust in the different participants of the social system, more specifically in their strong trust, as participants are generally publicly committed to the responsibilities assigned to the role they play in the social system. For example, the institutional-trust of a person in the judiciary system is grounded on her trust in the judge in charge of a trial, her trust in the attorney representing the plaintiffs, her trust in the investigator responsible for conducting an investigation, among other possibilities. 

\subsection{Quantitative Perspective of Trust}
\label{sec:quantitative-perspective}

Trust can increase or decrease in time, and we can trust certain individuals more than others. Also, the level of trust can be different for the same trustor, trustee and trustor intention, at the same time, depending on the factors that influence trust, which may characterize different contexts (we discuss the role of influences in context characterization later in section \ref{sec:influences}). 

To account for these scenarios, ONTrust assumes that trust can be quantified, even if it does not commit to any particular scale or measurement strategy. We begin the representation of the quantitative perspective of \textsf{Trust} by means of the \textsf{Trust Degree} (\textit{quality}) that inheres in the \textsf{Trust} entity. In UFO, a quality is an objectification of a property that can be directly evaluated (projected) into certain value spaces \cite{guizzardi2005ontological}. Common examples include a person`s weight, which can be measured in kilograms or pounds, and the color of a flower, which can be specified in the RGB or HSV color models. Thus, representing \textsf{Trust Degree} as a quality means that it can also be measured according to a given scale, such as a simple discrete scale like \textless{}Low,Medium,High\textgreater{} or a continuous scale (e.g. from 0.0 to 100.0).

In general, the trustor beliefs are not black and white, in the sense that a trustor needs to believe that a trustee either has a certain \textsf{Disposition} (or \textsf{Intention}) or not. In fact, they have an intrinsic quality that corresponds to the strength of a trustor belief \cite{jacquette2013belief}. For instance, it may be the case that ``Alice believes more strongly that Burger King is capable of making a good hamburger than it is capable of delivering orders on time''. Another example, which considers the same capability and different trustees, is ``the mother believes more strongly that her husband is capable of lifting a heavy object than their seven-year son''. Note that in this case, we are not comparing the performance of the father with that of his son when lifting a heavy object, because the child may not even be able to lift it. Nevertheless, performance levels are an important aspect to be considered with respect to capability beliefs. For example, a project manager may believe that both a junior and a senior analysts are capable of performing a particular task. However, he probably believes that the senior analyst is able to perform the task with a higher level of performance than the junior analyst. Castelfranchi and Falcone \cite{castelfranchi2010trust} claim that the degree of trust is a function of (i) the estimated quantitative level of the trustee’s quality on which the positive expectation is based and (ii) how much the trustor is sure of her evaluation about the trustee’s quality. In our approach, the \textsf{Belief Intensity} and the \textsf{Performance Level} are analogous to, respectively, (ii) and (i) in Castelfranchi and Falcone's proposal \cite{castelfranchi2010trust}.

Finally, another important aspect to be considered, related to beliefs about trustee dispositions, is how strongly the trustor believes a disposition may be manifested through the occurrence of certain events. For example: ``although Charlie believes that he can get a flat tire during a trip (which corresponds to a vulnerability belief about his car), he believes that the likelihood of this happening is very small''.

In the light of the above, our position is that trust is gradable and so are the beliefs that compose trust. Furthermore, when it comes to capability beliefs, the trustee's capabilities are gradable as well. For example, a supervisor may trust a PhD student to deliver a speech to an audience of twenty people, but not for an audience of one hundred people. In this case, regarding the PhD student’s capability of delivering a speech, the supervisor believes his performance level is higher for smaller audiences. Thus, we have that the supervisor’s belief is gradable and, consequently, his trust in the PhD student is gradable as well. Surprisingly, it may be the case that the supervisor is mistaken with respect to his belief about the PhD student’s capability, and in reality the PhD student’s performance level is the same no matter the size of the audience.

\begin{figure*}
\setlength{\fboxsep}{0pt}%
\setlength{\fboxrule}{0pt}%
\begin{center}
    \centering
    \includegraphics[width=\textwidth]{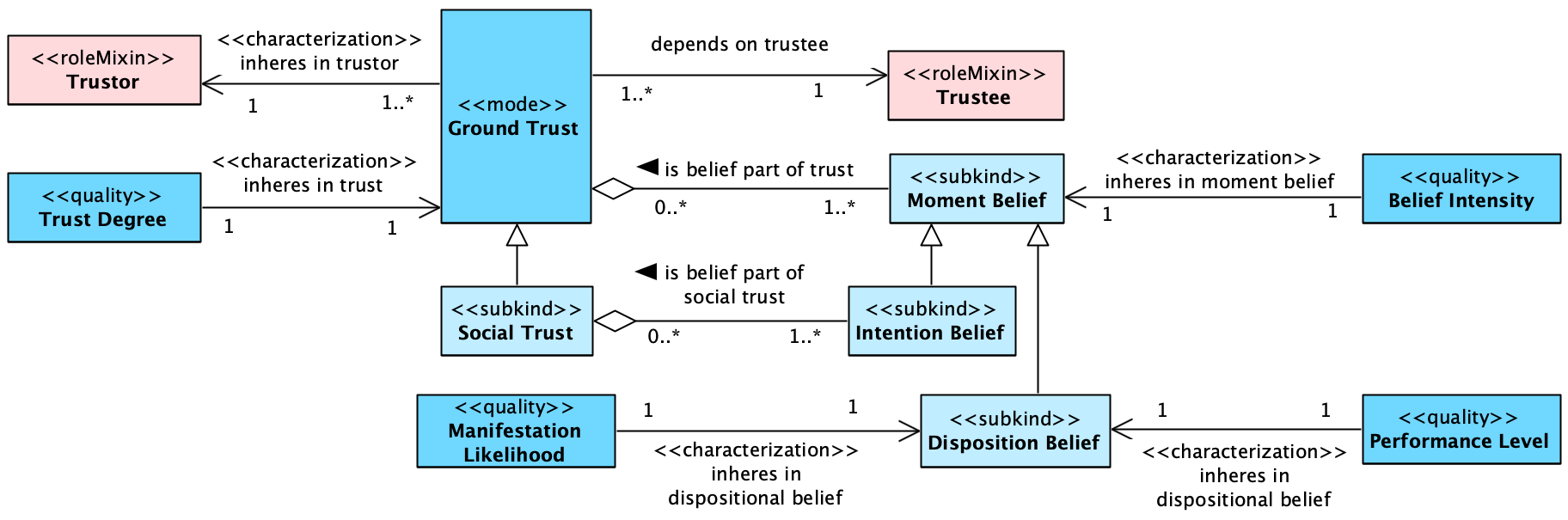}
    \caption{Quantitative Perspective of Trust}
    \label{fig:quantitative-perspective}
    \end{center}
\end{figure*}

Based on these considerations, we propose that the above-mentioned measures (belief intensity, performance level, and manifestation likelihood) be considered when quantifying trust. To account for the quantitative perspective of beliefs, we modeled these three belief-related measures (Figure \ref{fig:quantitative-perspective}), representing them as qualities that inhere in  \textsf{Moment Beliefs} (\textsf{Belief Intensity}) and \textsf{Disposition Beliefs} (\textsf{Performance Level} and \textsf{Manifestation Likelihood}). This means that they can be mapped into quality spaces, such as a discrete scale like  \textless{}Low,Medium,High\textgreater{} or a continuous one like \textless{}0-100\textgreater{} \cite{agudo2008model,ennew2007measuring,marsh1994formalising}. 

\subsection{Influences}
\label{sec:influences}

Several factors that influence trust have been discussed in the literature \cite{mayer1995integrative}. For instance, Castelfranchi and Falcone \cite{castelfranchi2010trust} argue that ``trust changes with experience, with the modification of the different sources it is based on, with the emotional or rational state of the trustor, with the modification of the environment in which the trustee is supposed to perform, and so on''. They claim that trust is a dynamic entity because it depends on dynamic phenomena. 

In this section, we categorize influence relations according to the ontological nature of the factors that characterize them. These categories are: (F1) trust influencing trust; (F2) mental biases; (F3) trust calibration signals; and (F4) trustworthiness evidence.

\subsubsection{F1: Trust Influencing Trust}

This category represents the situation in which \textsf{Trust} is influenced by another trust relationship. According to Castelfranchi and Falconi \cite{castelfranchi2010trust} ``in the same situation trust is influenced by trust in several rather complex ways’’. In the same work they also discuss the phenomenon of trust creating reciprocal trust, as well as how trust relations can influence each other. In fact, countless examples can be found in real life about trust influencing trust, either positively or negatively. For instance, citizens’ trust in the central bank positively influence their trust in the national currency. People’s trust in the healthcare system, in the experts defining vaccination strategies, and more generally in government bodies influence their trust in vaccines. 

 McKnight and Chervany \cite{mcknight2001trust} argues that \textsf{Institution-based Trust} affects \textsf{Social Trust} by making the \textsf{Trustor} feel more comfortable about trusting others in a given situation. For example, regulations and institutions may enable people to trust each other not because they know each other personally, but because licensing, auditing, laws or governmental enforcement bodies are in place to make sure the other person is either afraid to harm them or punished if they do so. This influence may also hold in the opposite direction. \textsf{Social Trust} may influence \textsf{Institution-based Trust} by generating positive beliefs about established social systems. For example, one's trust in the local police officer may increase one's trust in the ``judiciary system''.

\subsubsection{F2: Mental Biases}

This category represents situations in which \textsf{Trust} is influenced by \textit{mental moments} \cite{guizzardi2005ontological}. \textsf{Mental Moments} refer to the capacity of some properties of certain individuals to refer to possible situations of reality (see section \ref{sec:ufo}). A \textsf{Mental Moment} is existentially dependent on a particular \textsf{Agent}, being an inseparable part of its mental state. Examples of \textsf{Mental Moments} include \textsf{Perceptions}, \textsf{Beliefs}, \textsf{Desires} and \textsf{Intentions}. Also emotions (such as happiness, sadness and love) are existentially dependent entities (moments in UFO, see \ref{sec:ufo}), which can have moments themselves (such as duration and intensity) and therefore can be considered a special type of \textsf{Mental Moment}. The notion of mental moment is defined and discussed in section \ref{sec:ufo}. For an extensive discussion of this subject, refer to \cite{guizzardi2008grounding}.

\textsf{Mental Moments} can significantly influence \textsf{Trust}. Let us consider the example of a person who really wants to travel but does not have enough money to pay for a trip. One day she receives an email containing an unbelievable offer for an exotic destination at a very low price that is just about to expire. Although many will immediately think it is a scam, the person’s \textbf{desire} to travel may influence her to trust the email offer \cite{fischer2013individuals}. Another example would be people who are strongly committed to environmental preservation and tend to trust companies that support environmental sustainability \cite{chen2012enhance} due to their \textbf{beliefs}. Another one is the case of people who opposes vaccines, in times of COVID-19: their anti-vaccination \textbf{beliefs} negatively influence their trust in a COVID-19 vaccine. Still connected to the pandemic, the event of the US President Joe Biden receiving his first dose of COVID-19 vaccine on live television \cite{cnn2020binden} was \textbf{perceived} by many people in a positive way, thus positively influencing their trust in the vaccine. There is also the case of \textbf{beliefs} not related to specific trustees. An example discussed by McKnight and Chervany \cite{mcknight2001trust} suggests that some religious beliefs, which prescribe honesty and mutual love, lead people to generally assume that others are usually honest, benevolent, competent, and predictable. Regarding the influence of emotions, according to Dunn and Schweitzer \cite{dunn2005feeling}, for example, emotions with positive valence (such as happiness and gratitude) positively influence increase trust, while emotions with negative valence (like anger) negatively influence trust. Another example is the case of people that decide whether they can initially trust someone with respect to a particular goal, by examining the feelings that they have toward that person.

Another important aspect is the occurrence of events that can affect one's perception regarding a trustee. McKnight et al. \cite{mcknight2012events} discuss how trust changes in response to external events and propose a model that addresses the mental mechanisms people use as they are confronted by trust-related events, which ``indicates that trust may be sticky or resistant to change, but that change can and will occur'' \cite{mcknight2012events}. Castelfranchi and Falcone \cite{falcone2004trust} claim that the success of an action performed by the trustee in order to reach a goal of the trustor depends not only on the trustee’s capabilities but also on external conditions that allow or inhibit the realization of the task. To illustrate this point, the authors use the case of a violinist that will give a concert in an open environment. In general, people trust the violinist to play well. However, if it is particularly cold during the concert, their trust will decrease if they infer that the cold can hinder her ability to play. Similarly, in financial systems, the emergence of detrimental information about a financial agent can negatively affect public trust in this agent, which can lead to considerable adverse effects on one or several other financial institutions that can ultimately propagate to the entire financial system.

\subsubsection{F3: Trust Calibration Signals}

According to Riegelsberger et al. \cite{riegelsberger2005mechanics}, one of the ingredients for building sustainable trust is the emission of \textit{trust-warranting signals}, that is, signals that indicate trustworthy behavior of a trustee. In trust relations, once the trustee’s capabilities and vulnerabilities related to the beliefs of the trustor are known, it is possible to reason about the signals that the trustee should emit to indicate that it can successfully realize its capabilities and prevent its vulnerabilities from being manifested. These signals are specifically created to indicate a trustworthy behavior on the part of the trustee and therefore can influence trust. For example, information about how privacy and security measures are implemented could be provided as signals of the trustworthiness of a system.
Another example of signal emitted to reinforce trust is the establishment of a universal brand to create visual identity, so that users can identify the system interface elements in a clear and unambiguous way, thus facilitating the understanding and adoption of its functionalities.

Equally important are \textit{uncertainty signals}, i.e. signals that communicate uncertainties regarding the realization of capabilities and the prevention of vulnerabilities. Some examples are the publication of uncertainties about the accuracy of scientific findings, patient communication of uncertainties on the precision of medical diagnosis, investor communication of uncertainties in forecasting financial investments returns, communication to the public about uncertainties regarding the efficacy of vaccines, among others. While trust-warranting signals contribute to trust building, uncertainty signals allow trustors to adjust their trust level appropriately to the trustee’s trustworthiness, thus avoiding misplaced levels of trust. Research show that communicating uncertainty can be beneficial for maintaining trust and commitment over time \cite{batteux2021negative,tomsett2020rapid}. This is because building trust that is higher than the actual trustworthiness of the trustee might set trustors’ expectations too high, which may result in disappointment sooner or later. 

To model \textsf{Trust Calibration Signals} in ONTrust, we rely on concepts defined in the Communications Ontology proposed by Morales-Ramirez at al. \cite{morales2015ontology}, which was developed as an extension of the Unified Foundational Ontology. The Communication Ontology considers \textit{sender} and \textit{receiver} the two types of agents that are important in the communication process. The \textit{sender} is an agent that sends a \textit{message} through a \textit{speech act} \cite{searle1995construction}. According to Searle \cite{searle1995construction}, when a person (the \textit{sender}) says something she attempts to communicate certain things to the addressee (the \textit{receiver}), which affect either their beliefs and/or their behavior. In UFO-C, speech acts are defined as \textsf{Communicative Acts} (see section \ref{sec:ufo}) ---  actions that carry out the information exchanged in the communication process.

\begin{figure*}
\setlength{\fboxsep}{0pt}%
\setlength{\fboxrule}{0pt}%
\begin{center}
    \centering
    \includegraphics[width=\textwidth]{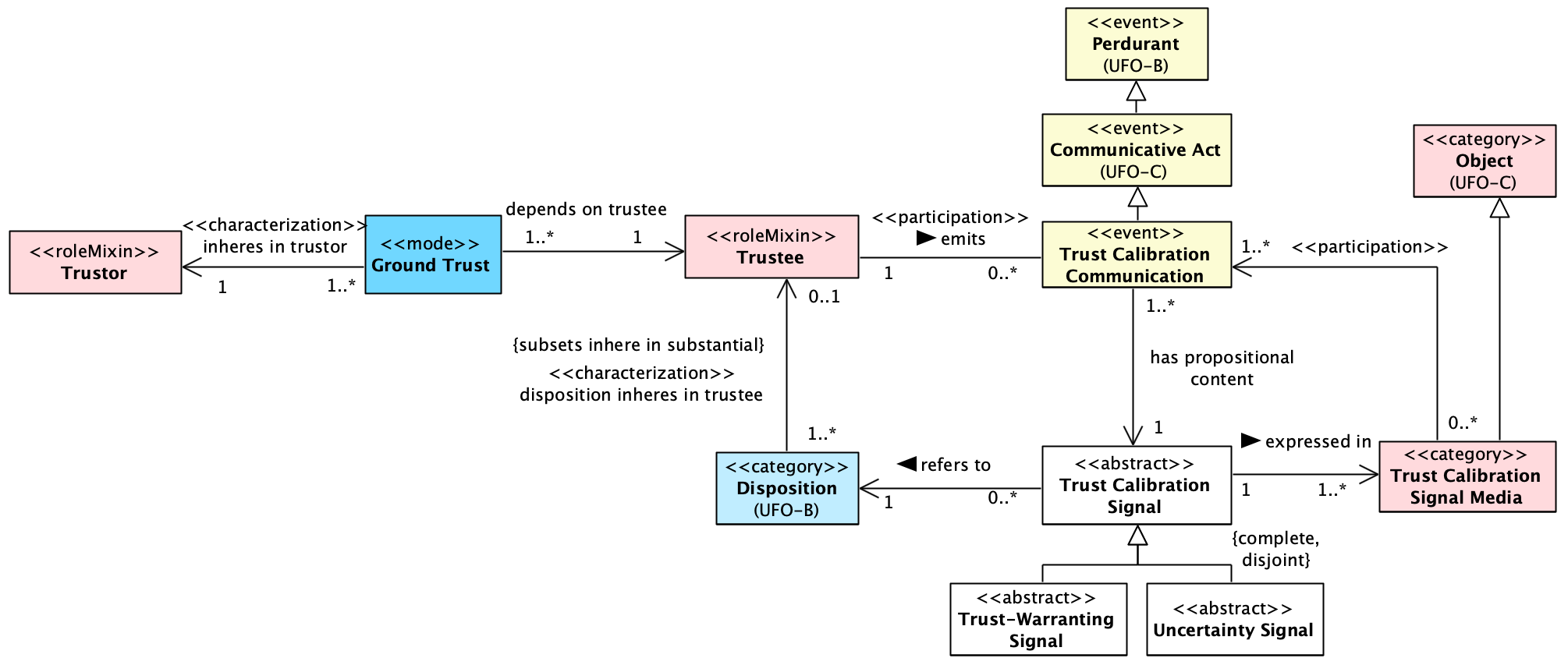}
    \caption{Trust Calibration Signals}
    \label{fig:trust-calibration-signal}
    \end{center}
\end{figure*}

As illustrated in Figure \ref{fig:trust-calibration-signal}, \textsf{Trust Calibration Signal} (the \textit{message}) is the propositional content of a \textsf{Trust Calibration Communication}, a \textsf{Communicative Act} that the \textsf{Trustee} (the \textit{sender}) emits to communicate either its trustworthiness --- in the case of \textsf{Trust-warranting Signals} --- or uncertainties in the realization of its capabilities, so that \textsf{Trustors} can adjust their trust levels --- in the case of \textsf{Uncertainty Signals}. The propositional content of this act (the \textsf{Trust Calibration Signal}) refers to a \textsf{Disposition} of the \textsf{Trustee}. 
The message (\textsf{Trust Calibration Signal}) may be expressed in a communication media (\textsf{Trust Calibration Media}) that is the instrument used to carry out communication. Examples of \textsf{Trust Calibration Media} are visual identity elements like uniforms and logos.

\subsubsection{F4: Trustworthiness Evidence}

Another trust influencing factor corresponds to \textit{trustworthiness evidence}, pieces of evidence that suggest that a trustee should be trusted. Similarly to trust-warranting signals, they suggest that a trustee can realize its capabilities and shield its vulnerabilities. However, differently from signals, which are purposefully emitted to suggest trustworthiness, evidence result from trustees’ trustworthy actions. Examples include: 

\begin{itemize}
    \item third-party certifications and credentials (e.g. John's TOEFL certification makes me believe that he can speak English, because I trust the certificate issued by a certain authority);
    
    \item performance history (e.g. accuracy of a medical diagnosis system); 
    
    \item track record (e.g. reviews from service recipients and statistics on its experience);  
    
    \item recommendations (e.g. my brother trusts a car mechanic and recommends his services to me); 
    
    \item reputation records (e.g. positive evaluations received by an Uber driver); 
    
    \item availability (e.g. a medical doctor you rarely succeed to make an appointment with is not trustworthy);  
    
    \item past successful experiences (e.g. all the products I purchased at Amazon arrived on time and in perfect condition);  
    
    \item transparency (e.g. offering information on what an artificial intelligence system is doing, as well as rationale for its decisions (aka explainability));  
    
    \item longevity (e.g. indication that a vendor has been in the market for a long time and that it is interested in continued business relationship with the client); and
    
    \item risk mitigation measures, which indicate that one is actively trying to prevent the manifestation of one's vulnerabilities. 

\end{itemize}

\begin{figure*}
\setlength{\fboxsep}{0pt}%
\setlength{\fboxrule}{0pt}%
\begin{center}
    \centering
    \includegraphics[width=\textwidth]{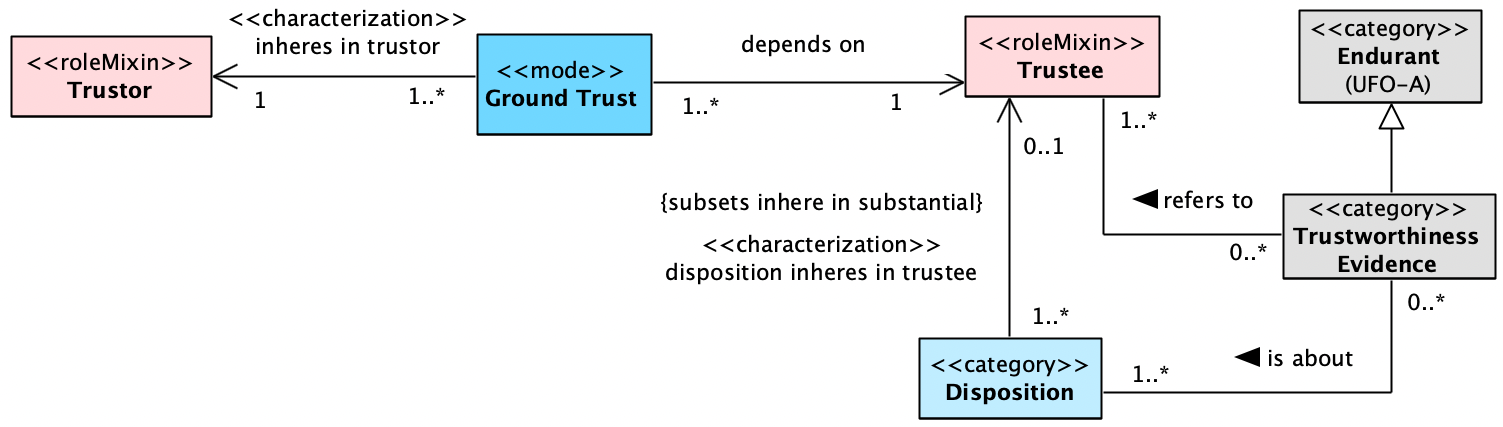}
    \caption{Trustworthiness Evidence}
    \label{fig:trustworthiness-evidence}
    \end{center}
\end{figure*}

Ontologically speaking, a piece of \textsf{Trustworthiness Evidence} is a \textit{social thing}, typically a \textit{social relator} (see section \ref{sec:ufo}) (e.g. a relator binding the certifying entity, the certified entity and referring to a capability, vulnerability, etc.), but also documents (\textit{social objects} themselves) that represent these \textit{social entities} (e.g., in the way a marriage certificate documents a marriage as a social relator). As illustrated in Figure \ref{fig:trustworthiness-evidence} we modeled \textsf{Trustworthiness Evidences} as roles played by \textit{endurants} (objects, relators, etc.)  related to a \textsf{Disposition} of the \textsf{Trustee}\footnote{Playing of the ``role" of \textsf{Trustworthiness Evidence} for a particular focal disposition is dependent on the belief of trustors, whose propositional content makes that connection between that player and that disposition. The \textsf{is about} relation in this model is, thus, derived from the propositional content of that belief. The \textsf{refers to} relation connected to the trustee is derived from the relation between that focal disposition and its bearer.}.

\begin{figure*}
\setlength{\fboxsep}{0pt}%
\setlength{\fboxrule}{0pt}%
\begin{center}
    \centering
    \includegraphics[width=\textwidth]{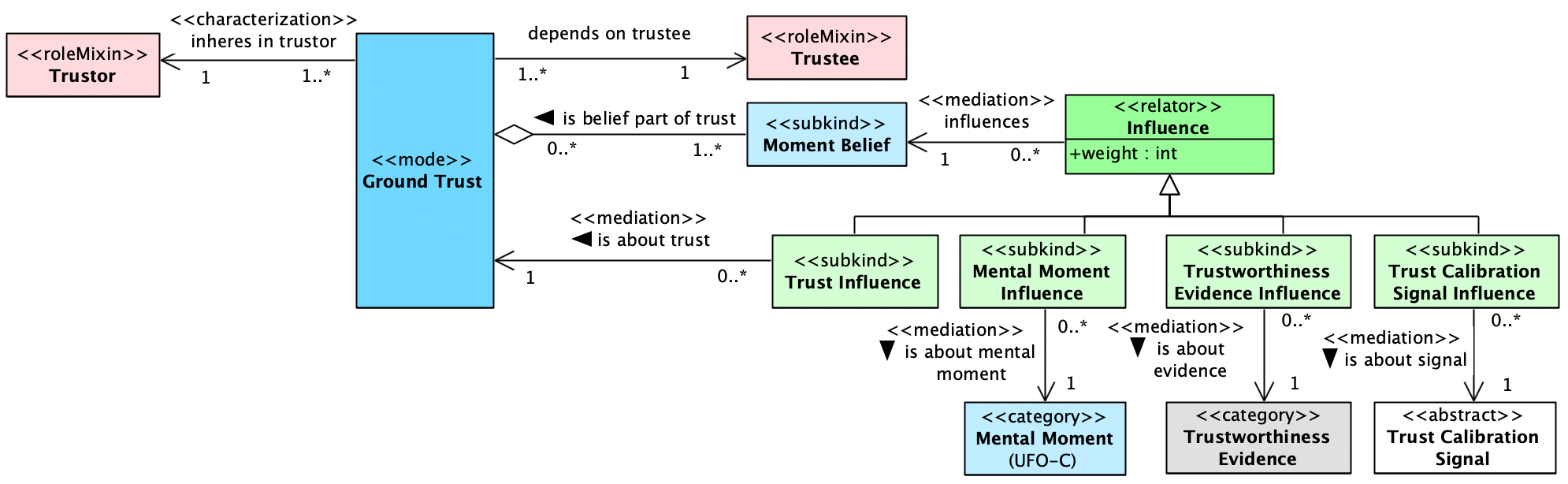}
    \caption{Influences}
    \label{fig:influence}
    \end{center}
\end{figure*}

To represent the role of influences, we included the \textsf{Influence} relator, which connects the sources of influence to the \textsf{Moment Beliefs} of the \textsf{Trustor} under their influence (Figure \ref{fig:influence}). We distinguish \textsf{Influence} according to the source of influence into: (i) \textsf{Trust Influence}, associated to a \textsf{Trust} entity (F1); (ii) \textsf{Mental Moment Influence}, associated to a \textsf{Mental Moment} (F2); (iii) \textsf{Trust Calibration Signal Influence}, associated to a \textsf{Trust Calibration Signal} (F3); and (iv) \textsf{Trustworthiness Evidence Influence}, associated to a \textsf{Trustworthiness Evidence} (F4). The property \textsf{weight} corresponds to the weight of an influence over a particular belief, as certain influences may weight more heavily than others.

\begin{figure*}
\setlength{\fboxsep}{0pt}%
\setlength{\fboxrule}{0pt}%
\begin{center}
    \centering
    \includegraphics[width=\textwidth]{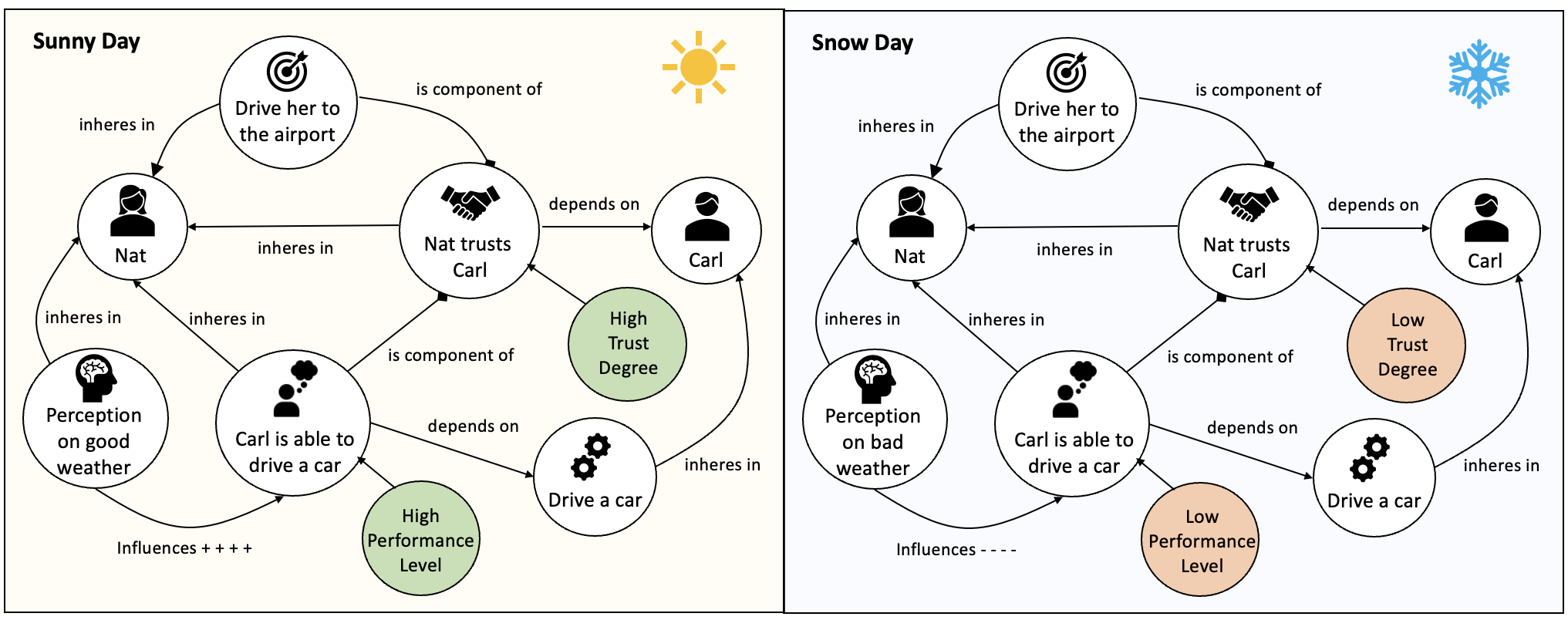}
    \caption{Trust depends on the context - Example}
    \label{fig:context}
    \end{center}
\end{figure*}

As previously mentioned in section \ref{sec:quantitative-perspective}, trust is a highly dynamic entity \cite{castelfranchi2010trust}. A trustor may trust a trustee for a certain goal in a given context, but not do so for the same goal in a different context. For example, Nat may trust Carl to drive her to the airport in a sunny day, but she does not trust him to do so when it is snowing (Figure \ref{fig:context}). 

Note that the factors that influence trust play a key role in context characterization. This is because changes in the environment may affect trustors’ perceptions of the world as well as their emotions, beliefs, and intentions. Such mental moments can influence the beliefs that compose the trust mental state of the trustor, thus leading to different trust levels. Also, the propositional content of the intention that trust is about takes part in context characterization. Considering the example mentioned in section \ref{sec:quantitative-perspective}, in which a supervisor trusts a PhD student to deliver a speech for a small audience, but not for a huge one, the propositional content of the supervisor’s intention may characterize different contexts and, consequently, bring about different levels of trust. Here we assume the context as the ``knowledge base’’ formed by the trust intention plus all those sources of influence along with the underlying propositional contexts, if applicable. Figure \ref{fig:context} illustrates how different contexts can lead to different levels of trust. In the left part of the figure, we have that Nat’s perception about the good weather positively influences her belief about Carl’s capability of driving a car. In this case, she believes that Carl is able to drive the car with a high-performance level, which leads to a high trust degree. Conversely, in the right part of the figure, her perception about the bad whether negatively influence her belief about Carl’s capability, which in the end results in a low trust degree.

\subsection{The Emergence of Risk from Trust Relations}

In ONTRust, the relation between trust and risk is analyzed based on the Common Ontology of Value and Risk \cite{sales2018cover}. COVER proposes an ontological analysis of notions such as Risk, Risk Event (Threat Event, Loss Event) and Vulnerability, among others.  Given the objectives of this paper, we focus here on the perspective of risk as a chain of events that impacts on an agent's goals, which the authors named Risk Experience. Risk Experiences focus on unwanted events that have the potential of causing losses and are composed by events of two types, namely threat and loss events. A \textsf{Threat Event} is the one with the potential of causing a loss, which might be intentional or unintentional. A \textsf{Threat Event} might be the manifestation of a \textsf{Vulnerability} (a special type of disposition whose manifestation constitutes a loss or can potentially cause a loss from the perspective of a stakeholder) or a threatening \textsf{Capability} 
(capabilities are usually perceived as beneficial, as they enable the manifestation of events desired by an
agent. However, when the manifestation of a capability enables undesired events that threaten agent’s abilities to achieve a goal, it can be seen as a threatening capability). The second mandatory component of a Risk Experience is a \textsf{Loss Event}, which necessarily impact intentions in a negative way (captured by a \textsf{hurts} relation between \textsf{Loss Event} and \textsf{Intention}) \cite{sales2018cover}.

\begin{figure*}
\setlength{\fboxsep}{0pt}%
\setlength{\fboxrule}{0pt}%
\begin{center}
    \centering
    \includegraphics[width=\textwidth]{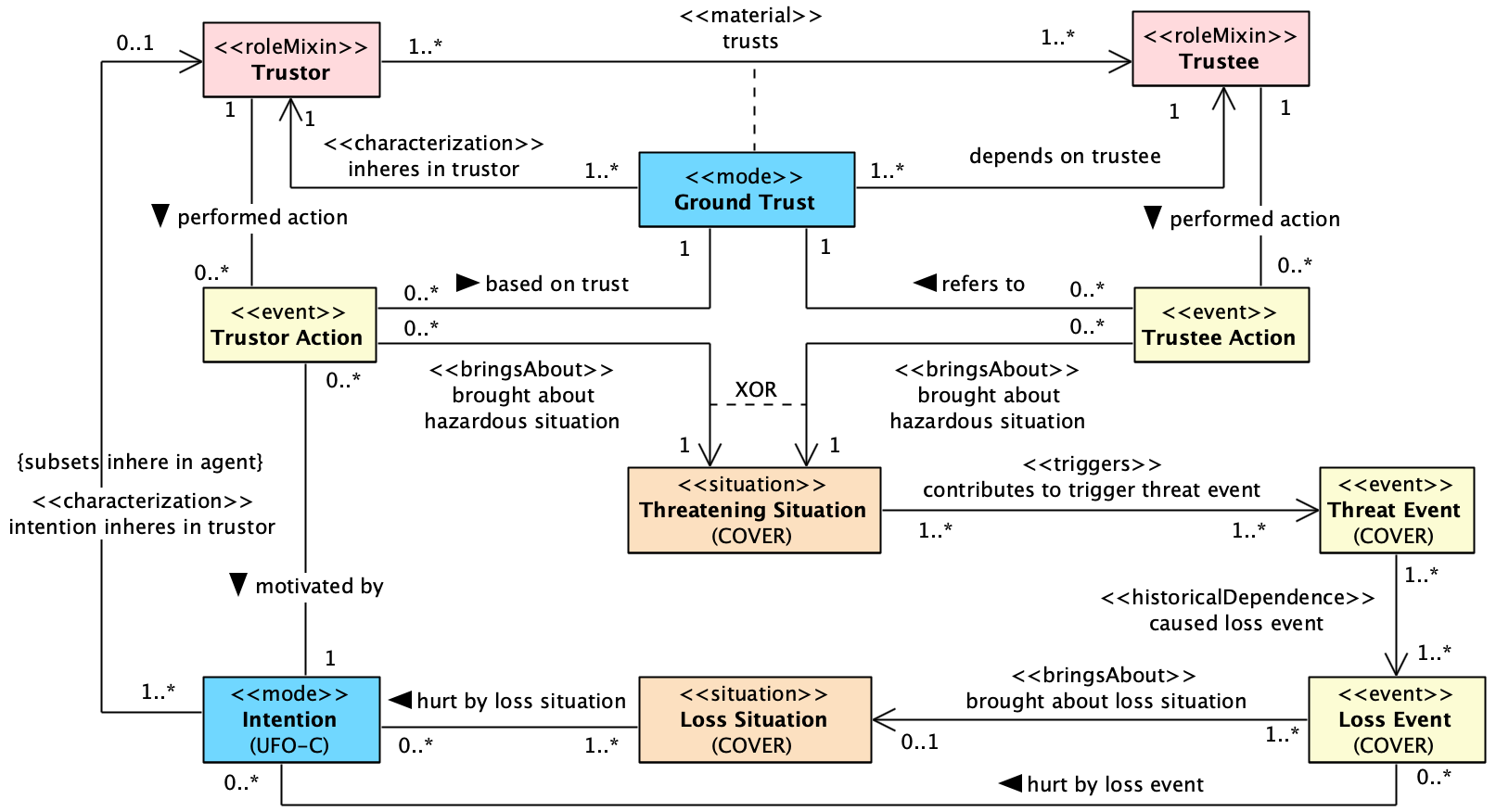}
    \caption{The Emergence of Risk from Trust Relations}
    \label{fig:trust-risk}
    \end{center}
\end{figure*}

The relation between trust and risk, together with its embedded concepts, is represented in the OntoUML model depicted in Figure \ref{fig:trust-risk}. As part of the behavioral perspective of trust, the \textsf{Trustor} may take some \textsf{Actions}, motivated by her \textsf{Intentions} and based on her \textsf{Trust} in the \textsf{Trustee}. These \textsf{Actions} may involve the \textsf{Trustee} or not (some examples are cooperation, information sharing, informal agreements, decreasing controls, accepting influence, granting autonomy, and transacting business \cite{mcknight2001trust}), however they are taken considering that the \textsf{Trustee} will behave according to the \textsf{Trustor}'s \textsf{Beliefs}. As previous mentioned, a \textsf{Trustor} may \textit{trust} in a \textsf{Trustee} but not take any \textsf{Action} based on this \textsf{Trust}. For this reason, the relationship between \textsf{Trust} and the \textsf{Trustor's Actions} is optional. 

Similarly, the \textsf{Trustee} may take some \textsf{Actions} aiming at performing \textsf{Capabilities} or preventing the manifestation of \textsf{Vulnerabilities}, as expected by the \textsf{Trustor}.

\textsf{Actions} performed by either the \textsf{Trustor} (based on her \textsf{Trust} in the \textsf{Trustee}) or the \textsf{Trustee} (regarding a \textsf{Trust} relation) brings about a \textsf{Resulting Situation}, which may satisfy the \textsf{Trustor} goals (and in this case it is considered a \textsf{Successful Situation}) or, in the worst case, may not have the desired result and the \textsf{Trustor} will not be able to achieve her goal. In this case, the \textsf{Resulting Situation} stands for a \textsf{Threatening Situation} that may trigger a \textsf{Threat Event}, which may cause a loss. The \textsf{Loss Event} is a \textsf{Risk Event} that impacts intentions in a negative way, as it hurts the \textsf{Trustor}'s \textsf{Intentions} of reaching a specific goal.

\subsection{ONTrust Axioms}
\label{sec:axioms}

This section presents the axioms in OCL (Object Constraint Language) \cite{omg2003ocl} that reflect the relevant constraints that are not directly implied by the previous models. This axiomatization was a result of a formalization process employing a iterative model simulation approach using the Alloy Analyzer \cite{benevides2010validating}. Briefly speaking, this approach consists, basically, in transforming OntoUML models into Alloy \footnote{Alloy specifications are used as input to Alloy Analyzer 4.2 tool, which generates instances of the model and represents these instances in a graphical representation.}  specifications and generating conforming instantiations of the model automatically. Such automatically generated model instantiations are then examined manually, to decide whether they are in conformance with a particular conceptualization.  If not, the OntoUML model is changed. Consistence of the axiomatization and OntoUML models is guaranteed by checking the satisfiability of the corresponding Alloy specification.

During model simulation, instantiations of ONTrust were generated and analyzed, and changes were made in the models for avoiding invalid instantiations. When necessary, OCL constraints were added to the ONTrust models for guaranteeing a certain rigor, and avoid invalid instantiations. 

\begin{figure*}
\setlength{\fboxsep}{0pt}%
\setlength{\fboxrule}{0pt}%
\begin{center}
    \centering
    \includegraphics[width=0.8\textwidth]{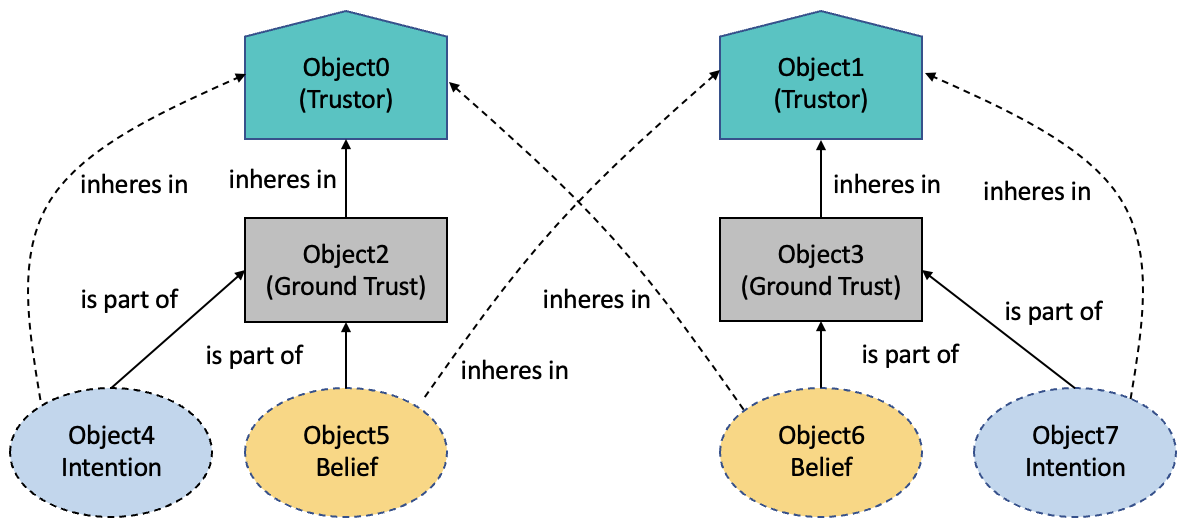}
    \caption{A possible instantiation of ONTrust, exemplifying an open instance cycle}
    \label{fig:open-cycle}
    \end{center}
\end{figure*}

\begin{figure*}
\setlength{\fboxsep}{0pt}%
\setlength{\fboxrule}{0pt}%
\begin{center}
    \centering
    \includegraphics[width=0.8\textwidth]{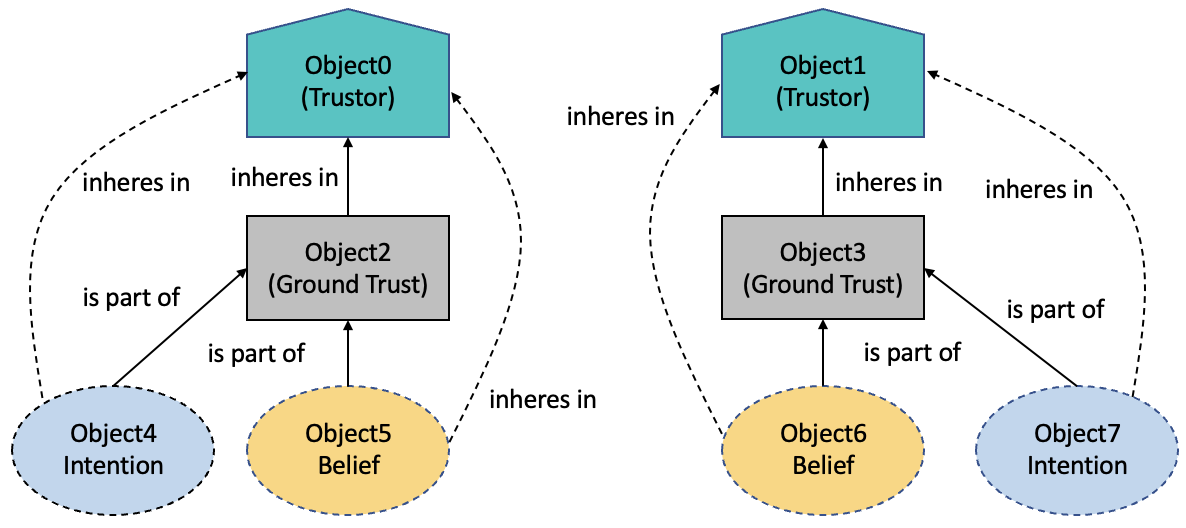}
    \caption{A possible instantiation of ONTrust, exemplifying a closed instance cycle}
    \label{fig:closed-cycle}
    \end{center}
\end{figure*}

A recurrent error-prone modeling decision identified during the simulations was the so-called Association Cycle (AssCyc) anti-pattern \cite{sales2015ontological}. The AssCyc anti-pattern occurs when an arbitrary number of classes are connected through the same number of relations in a way that composes a cycle. In other words, one can start navigating relations from any class in the cycle and arrive back to the starting point without going through the same relation and visiting the same class more than once (except for the first/last node). For example, in Figure \ref{fig:ground-trust}, the relations state that \textsf{Ground Trust} inheres in the \textsf{Trustor} and has as its parts an \textsf{Intention} and a set of \textsf{Beliefs} that inhere in the same \textsf{Trustor}. As revealed by the simulations, this structure allows for the instantiations of both open and closed cycles at the instance-level. Figure \ref{fig:open-cycle} presents a model instance that exemplifies an open cycle. Notice that \textsf{Ground Trust} is about an \textsf{Intention} that inheres in an \textsf{Agent} (\textsf{Trustor}) other than the one in which the \textsf{Belief} does. Alternatively, Figure \ref{fig:closed-cycle} presents an instantiation that characterizes a closed cycle. Note that, in this case, both the \textsf{Belief} and the \textsf{Intention} inhere in the same \textsf{Agent} (\textsf{Trustor}) as the \textsf{Ground Trust}.

In order to eliminate the unintended consequences induced by the presence of the AssCyc anti-pattern we applied a refactoring plan (extracted from Sales and Guizzardi's paper on ontological anti-patterns \cite{sales2015ontological}) that enforces the open cycle instantiation scenario at instance level through the specification of an OCL invariant. Table \ref{tab:refactoring-plan} presents this refactoring plan for the Ground Trust example mentioned above.

\begin{table}[h]
\centering
\caption{AssCyc anti-pattern refactoring plan and the \textsf{Ground Trust} example}
\label{tab:refactoring-plan}
\begin{tabular}{p{8cm}}

\noalign{\hrule height 0.8pt}
\rowcolor[HTML]{C0C0C0} 
\textbf{Ground Trust Example}                                  
\\ 
\addlinespace[0.8pt]
\rowcolor[HTML]{EFEFEF} 
The \textsf{Beliefs} and the \textsf{Intention} that are part of  a \textsf{Ground Trust}, must inhere in the same \textsf{Agent} (\textsf{Trustor}) as the \textsf{Ground Trust}.
\\ 

\noalign{\hrule height 0.8pt}
\rowcolor[HTML]{C0C0C0} 
\textbf{OCL Constraint}   
\\ 
\addlinespace[0.8pt]
\rowcolor[HTML]{EFEFEF} 
contex: Trustor \\
\rowcolor[HTML]{EFEFEF} 
inv : self.groundTrust.intention.trustor.asSet() \\
\rowcolor[HTML]{EFEFEF} 
-\textgreater includes(self)
\\ 
\noalign{\hrule height 0.8pt}

\end{tabular}
\end{table}

We present bellow the other cases in which the AssCyc anti-pattern was identified. For all of them, the refactoring plan presented in Table \ref{tab:refactoring-plan} was applied.

\begin{itemize}

\item \textsf{Complex Intentions} can only be composed of \textit{Intentions} that inhere in the same \textsf{Agent} (\textsf{Trustor}).

\item In an \textsf{Agreement} between a \textsf{Trustor} and a \textsf{Trustee}, all \textsf{Trustor Claims} and \textsf{Trustor Commintments} must inhere in the same \textsf{Agent} (\textsf{Trustor}). Similarly, all \textsf{Trustee Claims} and \textsf{Trustee Commintments} must inhere in the same \textsf{Agent} (\textsf{Trustee}).

 \item In \textsf{Weak Trust}, the \textsf{Social Commitment Belief} must inhere in the same \textsf{Agent} (\textsf{Trustor}) as the \textsf{Intention} and the other \textsf{Beliefs} that are part of that instance of \textsf{Weak Trust}.

  \item In \textsf{Strong Trust}, the Trustor's Claims and Commitments that compose the \textsf{Agreement} in which \textsf{Strong Trust} is grounded must inhere in the same \textsf{Agent} (\textsf{Trustor}) as the \textsf{Intention} and the other \textsf{Beliefs} that are part of that instance of \textsf{Strong Trust}. The same goes for the Trustee's Claims and Commitments, which must inhere in the same \textsf{Agent} (\textsf{Trustee}).
  
\item In \textsf{Trust Delegation}, the \textsf{Trustor} and \textsf{Trustee} must be the same as the ones in the \textsf{Weak Trust} in which that instance of \textsf{Trust Delegation} is grounded.

Finally, Table \ref{tab:ocl-xor} presents the OCL constraint created for the construct XOR in the model of Figure \ref{fig:trust-risk}. According to this model, a \textsf{Threatening Situation} can be brought about by either a \textsf{Trustor Action} or a \textsf{Trustee Action}. 

\begin{table}[h]
\centering
\caption{OCL constraint for the XOR construct}
\label{tab:ocl-xor}
\begin{tabular}{p{8cm}}

\noalign{\hrule height 0.8pt}
\rowcolor[HTML]{C0C0C0} 
\textbf{XOR}                                  
\\ 
\addlinespace[0.8pt]
\rowcolor[HTML]{EFEFEF} 
 A \textsf{Threatening} can be brought about by either a \textsf{Trustor Action} or a \textsf{Trustee Action}.
\\ 

\noalign{\hrule height 0.8pt}
\rowcolor[HTML]{C0C0C0} 
\textbf{OCL Constraint}   
\\ 
\addlinespace[0.8pt]
\rowcolor[HTML]{EFEFEF} 
context ThreateningSituation \\
\rowcolor[HTML]{EFEFEF} 
inv : self.TrustorAction-\textgreater size() + \\
\rowcolor[HTML]{EFEFEF} 
self.TrusteeAction-\textgreater size() = 1
\\ 
\noalign{\hrule height 0.8pt}

\end{tabular}
\end{table}

\end{itemize}

\section{Use Case Illustrations}
\label{sec:case-studies}

In this section, to illustrate the ontology expressivity and capacity to represent real-world situations, the Reference Ontology of Trust is applied to model two cases from the literature. First, in section \ref{sec:elections} we present the modeling of a case of trust in IT-mediated elections presented in the case study reported by of Avgerou \cite{avgerou2013explaining}. Then, in section \ref{sec:ai} we model a case on trust in artificial intelligence (AI) for medical diagnosis, reported by Juravle et al. \cite{juravle2020trust}.

\subsection{Trust in IT-Mediated Elections}
\label{sec:elections}

This section presents a realistic study in which ONTrust is applied to model a case of trust in IT-mediated elections in Brazil, reported in \cite{avgerou2013explaining}. In this work, the authors refer to electronic voting (or e-voting) as the use of electronic technology to register voters, perform voting, and produce the voting results. According to them, attempts to use electronic voting technologies have failed in several countries due to a lack of public trust in IT-mediated elections.  In this context, they analyze the electronic elections in Brazil, which stands out as one of the few cases of nationwide e-voting that appear to have been used without raising much public concern, disputes or controversies over election results. In their study, the authors seek to understand the social mechanisms that contributed to the perception of trustworthiness of the voting conduct.

We use a tabular approach that describes the instantiation of ONTrust concepts for this case, presented in Appendix 1. Such instances express the different types of trust relations involved in the Brazilian e-voting ecosystem, as well as their components, in terms of concepts and relations defined in the ontology. While Table \ref{tab:trust-types-table} concerns aspects related to the different types of trust relations, Table \ref{tab:trust-influences-table} refers to the factors that can influence trust.

Table \ref{tab:trust-types-table} presents the building blocks that compose electors' trust in e-voting in Brazil. In general, the process of e-voting involves two main categories of social actors: citizens in their voter role, and electoral authorities in their role of organizers and guarantors of the running of voting and production of voting results. In this case study, the authors consider the object of trust to be the composite entity of the organization responsible for the conduct of elections (electoral authorities), and the technologies used to register voters, record the casting of votes, aggregate and communicate the results. In this paper we use the term \textit{e-voting ecosystem} to refer to this composite entity. 

As presented in Table \ref{tab:trust-types-table}, the electors and the e-voting ecosystem play the roles of \textsf{Trustor} (concept \#1) and \textsf{Trustee} (concept \#2), respectively. Electors trust the e-voting ecosystem to choose their leaders, which correspond to their \textsf{Intention} (concept \#4). According to Avgerou \cite[p.~425]{avgerou2013explaining}, trust in e-voting concerns ``expectations and perceptions regarding the extent to which the voting part of election is organized and conducted lawfully and competently, and the extent to which the voting results are an accurate representation of citizens’ preferences''.

By analyzing the case in the light of ontology, it is possible to identify slight variations in the trust relations. We have \textsf{Ground Trust} (concept \#3), which includes the set of beliefs of the electors regarding dispositions --- capabilities and/or vulnerabilities --- of the e-voting ecosystem. In Table \ref{tab:trust-types-table}, these disposition beliefs correspond to concepts \textsf{Capability Belief} (\#5) and \textsf{Vulnerability Belief} (\#7). As made explicit in ONTrust, disposition beliefs refer to types of dispositions that electors believe the e-voting ecosystem has, which correspond to the concepts \textsf{Capability Type} (\#6) and \textsf{Vulnerability Type} (\#8). Furthermore, we have \textsf{Social Trust} (concept \#9), which includes the beliefs of electors about the intentions of the electoral authorities (a component part of the e-voting ecosystem and, consequently, a social agent trustee). \textsf{Intention Beliefs} and \textsf{Intention Types} correspond to concepts \#10 and \#11, respectively. In addition, our analysis revealed the existence of \textsf{Strong Trust} (concept \#12), as (i) the electors believe that the electoral authorities are publicly committed to conducting the e-voting according to the law, as well as the e-voting ecosystem as a whole is committed to providing accurate results (\textsf{Social Commitment Belief}, concept \#13) and (ii) the electoral authorities are publicly committed to conducting the e-voting according to the law (\textsf{Trustee Commitment}, concept \#14). Finally, we observe the occurrence of \textsf{Institution-based Trust} (concept \#15) that corresponds to electors’ trust in the Electoral Justice System (a social system) to support e-voting. Note that this instance of institutional-based trust specializes a new instance of ground trust (different from the one that corresponds to the trust of electors in the e-voting system), in which the electors and the Electoral Justice system are the trustor and trustee, respectively. In addition, as explained in section \ref{sec:institution-based-trust}, this institutional-based trust is grounded on the electors’ \textsf{Social Trust} in the participants of the Electoral Justice system, such as the Superior Electoral Court (TSE), the regional electoral courts (TRE’s), the judges and the electoral boards.

Table \ref{tab:trust-influences-table} presents the many factors that can positively (+) or negatively (-) influence the trust of electors in the e-voting ecosystem, namely, \textsf{Mental Moments} (concept \#17),  \textsf{Trust Calibration Signals} (concept \#18), \textsf{Trustworthiness Evidence} (concept \#19) and other \textsf{Trust Relations} (concept \#16). 

An interesting observation about these influences is regarding \textit{beliefs about types} (see section \ref{sec:ground-trust}): Brazilians’ negative beliefs about the honesty of politicians (a type) can negatively influence their trust in the e-voting ecosystem.

Note that once that the building blocks that compose trust, as well as the factors that can influence it, are known, it is possible to reason about what can go wrong with respect to any of these elements, which may hurt the intentions of the trustor. These unwanted events correspond to risk events for which mitigation strategies may be defined in advance. 

For example, by using ONTrust to model trust of electors in e-voting, we identified that some important positive influences reinforcing electors' trust are: (a) their perception of electoral authorities as competent and compromised with the well-functioning of the e-voting ecosystem; (b) voters' successful previous experiences with e-voting and with IT solutions in general; and (c) the the lack of dispute over results.

Regarding (a), let us consider the situation in which actions taken by electoral authorities raises doubts regarding their compliance with the law or their commitment to the security and proper functioning of e-voting. This kind of situation could negatively influence electors' trust, as they might fear that their goal of electing their leaders through fair elections would be jeopardized by one of these actions. Therefore, misconduct on the part of electoral authorities poses a threat to maintaining electors' trust.

With respect to (b), a possible malfunction of the e-voting system during the elections could negatively influence public confidence in technological solutions and, more specifically, in the e-voting ecosystem. Hence, the proper functioning of the e-voting system, as well as positive experiences with the use of IT, are key ingredients to maintaining trust.

Finally, regarding (c), another hypothetical situation that can pose risks to the maintenance of trust is political parties questioning the vote arrangements or challenging the election results, as these actions may influence public perceptions and beliefs about the trustworthiness of the e-voting ecosystem. For example, 2022 Presidential elections left a legacy of public concern about risks of fraud, amid a polarized presidential race, in which the far-right political parties have questioned the reliability of electronic voting machines. Although it has generated a lot of questioning, the negative influence of these allegations was apparently neutralized by the strong positive influence of other factors, such as trust in the Electoral Justice system, a strong media campaign in favour of the reliability of the e-voting ecosystem, the endorsement and acceptance of the election results by other political parties, as well as electors' past successful experiences with the electronic voting system.

\subsection{Trust in artificial intelligence for medical diagnosis}
\label{sec:ai}

In this section we apply ONTrust to model a case of \textit{trust in artificial intelligence for medical diagnosis} discussed by Juravle et al. \cite{juravle2020trust}. In this work, the authors conducted three online experiments to investigate how much trust people have in AI as a potential diagnosis tool in the medical healthcare field. Experiment 1 aimed at better understanding how patients’ trust in the diagnosis is affected when it is given by an AI doctor rather than by a human doctor, and whether this trust depends on severity of disease or familiarity with technology. Experiment 2 examined if patients expected higher standards of performance from AI, as compared to human doctors, in order to trust treatment recommendations. Finally, Experiment 3 investigated about possibilities to increase trust in AI. In this paper, we are particularly interested in using ONTrust to model the case of trust in AI for medical diagnosis investigated in Experiment 1, thus, in the rest of this section we refer only to this experiment. 

In a nutshell, Experiment 1 investigated:

\begin{itemize}
\item how trust in AI compares to trust in a human doctor when AI is used as a primary or secondary diagnosis tool, offering a second opinion to that of a human doctor in the latter case;
\item how disease risk impacts trust toward AI diagnosis. Diseases were classified as high-risk if they could be life threatening when left untreated, otherwise they were considered low-risk;
\item whether familiarity with technology results in greater trust for AI diagnosis.
\end{itemize}

During the experiment, all participants were presented with eight hypothetical medical scenarios in random
order, each followed by two proposed diagnosis (first and second). The scenarios varied on: disease risk (high or low), first diagnosis result (positive or negative), and second diagnosis result (confirming or dis-confirming the first diagnosis result). Depending on the source of the first and second diagnosis, participants were allocated to one of three groups: Human-AI, AI-Human, Human-Human. 
After each diagnosis, they were asked to indicate a percentage score (0–100\%) for how much they trusted the diagnosis. Additionally, participants in the first two groups were given a text explaining how AI is used for medical diagnosis. Familiarity with technology was measured with the Media and Technology Usage and Attitudes Scale \cite{rosen2013media}.

The main results reported by the authors for Experiment 1 are as follows:

\begin{itemize}

\item [R1] A significant main effect of doctor type was found on the trust ratings data, with higher trust in human, as compared to AI diagnosis.

\item [R2] A significant main effect of disease risk was evidenced on the ratings data, with participants placing higher trust in low-risk, as compared to high-risk diagnosis.

\item [R3] The main effect of doctor type, and the interaction effect between disease risk and doctor type, suggested that trust increased more when the second confirming opinion was provided by a human doctor vs an AI doctor, with this difference being larger for high-risk diseases.

\item  [R4] When adding familiarity with technology as a co-variate, only the main effect of doctor type remained significant, but not the main effect of disease risk. Familiarity with technology did not correlate significantly with trust in either low-risk or high-risk diseases diagnostic.

\item [R5] In summary, Experiment 1 provided evidence that people tend to trust human diagnosis more than AI diagnosis. Results indicate overall lower trust in AI, as well as for diagnosis of high-risk diseases. Participants trusted AI doctors less than humans for first diagnosis, and they were also less likely to trust a second opinion from an AI doctor for high risk diseases.
\end{itemize}

In the sequel, we instantiate the ontology with the analysis and results of Experiment 1.  Figures \ref{fig:trust-human} to \ref{fig:trust-institutional}, presented in Appendix 2,  show a graphical representation of the ontology instantiation. 

We adopt the following coding to refer to instances of key ONTrust concepts hereafter: \textbf{INT} for intention; \textbf{BEL} for belief; \textbf{CAP} for capability; \textbf{VUL} for vulnerability; \textbf{CMT} for commitment; \textbf{AGR} for agreement; \textbf{MMI} for mental moment influence; \textbf{TSI} for trust-warranting signal (trust calibration signal) influence; \textbf{TEI} for trustworthiness evidence influence; and \textbf{TRI} for trust relation influence.

In the trust case of Experiment 1, we have the patient playing the role of \textsf{Trustor} and both the human doctor and the AI doctor playing the roles of \textsf{Trustees}. For this reason, we split the ontology instantiation into the following trust relations: (i) patient trusts human doctor to diagnose a disease;  and (ii) patient trusts AI doctor to diagnose a disease. In both cases, \textit{diagnose a disease} (INT) corresponds to the \textsf{Intention} the trust relation is about. Figures \ref{fig:trust-human} and \ref{fig:trust-ai} show a graphical representation of the ontology instantiation for patients' trust when the trustee is, respectively, a human doctor and an AI doctor.

In Figure \ref{fig:trust-human}, we have that patients believe that the \textit{human doctor is able to accurately diagnose a disease} (BEL), which is related to the human doctor \textit{capability to diagnose diseases} (CAP). Patients also believe that the human doctor not only \textit{intends to accurately diagnose the disease} (BEL), but he is also \textit{committed to making accurate diagnosis} (BEL). The existence of an \textit{employment contract between the human doctor and the hospital} (AGR) represents an \textsf{agreement} that formalizes this \textsf{commitment} (CMT), and thus the trust relation between the patient and the human doctor has been classified as a case of \textsf{Strong Trust}.

Figure \ref{fig:trust-ai} shows that patients believe that the \textit{AI doctor is able to accurately diagnose a disease} (BEL), which is related to its \textit{capability to diagnose diseases} (CAP). Patients also believe that \textit{the AI doctor intends to accurately diagnose the disease} (BEL) --- an indirect intention as it refers to an intention of the hospital that provides the AI doctor services. The same goes for the patient's belief that \textit{the AI doctor is committed to making accurate diagnosis} (BEL) --- again, an indirect commitment. Note that in this case, as it cannot be said that there is clearly an agreement formalizing this commitment, the trust relation between the patient and the AI doctor has been classified as a case of \textsf{Weak Trust}.

Figure \ref{fig:trust-institutional} shows the existence of \textsf{Social Trust} between patients and the hospital. In this case, patients trust the hospital \textit{to receive good quality and reliable treatments} (INT), and they believe that \textit{the hospital is able to provide reliable healthcare services} (BEL).

Figures \ref{fig:quantitative-human} and \ref{fig:quantitative-ai} illustrate the quantitative perspective of trust for the human doctor and the AI doctor cases, respectively. As presented in Figure \ref{fig:quantitative-perspective}, \textsf{trust degree} and \textsf{performance level} \textsf{characterize} trust and beliefs of the trustor, respectively. For the sake of simplicity, the names of these relations were omitted in Figures \ref{fig:quantitative-human} and \ref{fig:quantitative-ai}.  As reported in the results (R4), people tend to trust human diagnosis more than AI diagnosis. For this reason, \textit{trust degree} is instantiated as ``high'' in Figure \ref{fig:quantitative-human} and ``medium'' in Figure \ref{fig:quantitative-ai}. Similarly, as the results indicate that people believe that human doctors are more capable than AI doctors, \textit{performance level} is instantiated as ``high'' in Figure \ref{fig:quantitative-human} and ``medium'' in Figure \ref{fig:quantitative-ai}.

Finally, Figures \ref{fig:influence-human} and \ref{fig:influence-ai} illustrate the factors that influence patients' trust in the human doctor and the AI doctor, respectively. As shown in Figure \ref{fig:influence} \textsf{trust relations}, \textsf{mental moments}, \textsf{trustworthiness evidence} and \textsf{trust calibration signals} may \textit{influence} beliefs of the trustor. For the sake of clarity the names of theses relations are not shown in Figures \ref{fig:influence-human} and \ref{fig:influence-ai}.

As previously mentioned, the experiment results revealed that \textit{disease severity} (MMI) can influence patients' trust in medical diagnosis, with patients placing higher trust in low-risk, as compared to high-risk diagnosis. They indicate that for diagnosing high-risk diseases their trust is greater in human doctors than in AI. Furthermore, people are less likely to trust a second opinion from an AI doctor for high-risk diseases. 

\textit{Successful track record of diagnosing diseases} (TEI)  is an example of \textsf{trustworthiness evidence} that can influence trust, both for the human doctor and the AI doctor.

\textit{Social trust of patients in the hospital} (TRI) is an example of \textsf{trust relation} that positively influences patients's trust in the AI medical diagnosis services provided by the hospital.

As revealed by the experiment results, trust can vary if AI is being used as a primary or secondary diagnosis tool, offering a second opinion to that of a human doctor. Therefore, the \textit{diagnose stage} (MMI) can influence patients' trust in medical diagnosis when the trustee is an AI doctor. 

Although \textit{technology familiarity} (MMI) is a source of influence that usually leads to greater trust in IT solutions (as indicated in the e-voting case previously presented), the hypothesis that it would result in greater trust for AI diagnosis could not be confirmed by the experiment results. 

Finally, \textit{providing explanations on how AI is used for medical diagnosis} (TSI) is an example of \textsf{trust calibration signal}, more specifically \textsf{trust-warranting signal}, that can influence trust in the AI doctor.

\section{Related Work}
\label{sec:related-work}

Several trust-modeling approaches have been proposed over the years. In the context of the semantic web and social networks, most approaches focused simply on the representation of trust relations. One example is the work of Golbeck at al. \cite{golbeck2003trust}, which proposes an extension of the Friend of a Friend (FOAF) ontology to allow users to state and represent their trust in individuals they know. This schema also allows users to indicate a level of trust for people they know, which can be considered close to the concept of trust degree in ONTrust. In addition, trust can be given in general, or limited to a specific topic (possibly having a connection with an intention of the trustor). Similar to ONTrust, in this proposal the trustee can be a person (a cognitive agent) or another subject. Although it does not model the factors that can influence trust, they suggest that network graph can be used to infer the trust that one user should have for individuals to whom they are not directly connected.

Another example is the Proof Markup Language Trust Ontology (PML-T) \cite{mcguinness2007pml}, which provides an extensible set of primitives for encoding trust information associated with information sources. PML-T was created as part of the Proof Markup Language, a standard developed by the Stanford University that defines primitive concepts and relations for representing knowledge provenance \cite{mcguinness2007pml}. It defines trust and belief relations involving a trustor, a trustee (the information source), and pieces of information. The belief relation shows the belief of an agent about the source. The trust relation shows an agent's overall beliefs about information from the specified source. PML-T also allows the representation of trust ratings (as either encoded by users or calculated by trust algorithms), which are similar to the concept of belief intensity in ONTrust. Although providing a framework for encoding trust relations, PML-T does not prescribe a way for representing trust itself. 

 Dokoohaki and Matskin \cite{dokoohaki2008effective} propose a trust ontology for the design of trust networks on semantic web-driven social systems. The main component of their ontology is the trust relation that represents the connection between entities on the network. Every relation has a set of main properties that describe its nature and purpose, such as a topic and a value that represents the trust level. The authors also define a set of auxiliary properties for the trust relation, such as a goal that stands for the reason for establishing the relation and a recommender, which is a person on the network that has recommended the trustee. Comparing to ONTrust, the concepts of goal property and recommender can be considered close in meaning to the concepts of trustor intention and trustworthiness evidence in ONTrust, respectively. Furthermore, the relation between trust and risk is not mentioned. 
 
Huang and Fox \cite{huang2006ontology} proposed a logical theory of trust in the form of an ontology that gives formal and explicit specification for the semantics of trust. The authors define two types of trust, namely, trust in belief and trust in performance. In the former, the trustor believes that something the trustee believes is true (for example: Mary wants to order a product and her friend John suggests she buys it from an online store he believes always delivers the orders on time. Mary does not know the online store at the time, but she believes what John believes, which is that the store delivers the orders on time). Although the authors name this concept trust belief, in our view, it corresponds to a belief rather than a trust relation. It can also be seen as a kind of trust influence. In the latter, the trustor believes in a piece of information created by the trustee or in the performance of an action committed by the trustee, both in a context within the trustor's context of trust. In this sense, trust in performance is similar to the concepts of ground trust in ONTrust. Other types of trust, trust influences, as well as the relation between trust and risk are not represented in Huang and Fox's ontology \cite{huang2006ontology}.

Viljanen \cite{viljanen2005towards} surveyed and classified thirteen computational trust models to create an ontology of trust. In his proposal, trust is represented as a relation between a trustor and a trustee, which depends on the action that the trustor is attempting and on the competence of the trustee. Additionally, Viljanen defines an element of confidence attached to the trust relationship, as well as a set of third party opinions in the form of reputation information. These can be seen as forms of of trustworthiness evidence in ONTrust. The author uses the concept of business value to represent both value and risk associated to the trustor's action. By attaching business values to the action, the ontology is able to represent the potential impact, positive or negative, of the action that the trustor is attempting. However, the representation of the relation between trust and risk lacks a more detailed description. For example, the ontology does not make it explicit how risk events are triggered, nor how they affect the trustor.

Secure Tropos \cite{giorgini2005modeling} is a security-oriented extension of the agent-oriented software development methodology Tropos \cite{bresciani2004tropos} that adds both security and trust as part of the software development process. In Secure Tropos the concepts of trust and delegation are combined to represent dependence relations between agents. Their constructs for trust refer to existent trustworthiness between actors along trust relations rather than specify the nature of the concept of trust. Secure Tropos differs from our approach not only regarding this particularity, but also because it does not elaborate on the factors that can influence trust nor on the close relation between trust and risk. Tropos models support a role-based approach to trust \cite{giorgini2005modeling}, in which the trustee is represented by roles or positions rather than by individual agents. It is based on the assignment of objectives and responsibilities to roles (or positions), and the assignment of agents to roles. According to this approach, an individual X should trust another individual Y due to the role played by Y, even if they have never met before. By analyzing this situation in the light of ONTrust, it possible to identify some correspondence to Strong Trust: X trust Y because X knows that Y is committed to the responsibilities assigned to Y’s role.

Riegelsberger et al. \cite{riegelsberger2005mechanics} propose a framework on the mechanics of trust, in which they identify contextual (temporal, social, and institutional embeddedness) and intrinsic (ability and motivation) properties that warrant trust in another actor, which they name trust-warranting properties. They also describe how the presence of these properties can be signaled. Their analysis focus on correctness of trust-decisions. According to them, signals of trust-warranting properties are the basis for trust. In their model, they identify two broad categories of signals: symbols and symptoms, which are analogous to the ONTrust concepts of trust-warranting signals and trustworthiness evidence, respectively. They also acknowledge on the existence of institutional-based trust and discuss its influence on other trust relations. Despite these similarities, their work differs from what we propose here, as they do not consider uncertainty signals and other factors that may influence trust, such as the mental state of the trustor. Also, they do not provide an ontological account for the concepts represented in their model.

Castelfranchi and Falcone \cite{castelfranchi2010trust} made an important contribution with their theory of trust. In their work, they investigate what kind of beliefs and goals are necessary for trust to formulate several necessary conditions, such as the trustor having a goal and the belief that the trustee is competent and willing to achieve this goal. They also consider a behavioral aspect of trust, which is related to the notion of acting on trust. In our proposal we rely largely on their theory to formalize the general concept of trust, as well as the concept of social trust. As for the institution-based trust, Castelfranchi and Falcone \cite{castelfranchi2010trust} state that it ``builds upon the existence of shared rules, regularities, conventional practices, etc. and relies on this, in an automatic, non-explicit, mindless way'', however the authors do not formalize this aspect of trust. Likewise, the relation between trust and risk is emphasized in their theory, but is not formalized. Finally, in the same work, they present trust dynamics in different aspects: (i) how trust changes on the basis of the trustor’s experiences, which is related to the ONTrust concepts of \textit{trustworthiness evidence} and \textit{influence}; (iii) how trust is influenced by trust; how diffuse trust diffuses trust (that is how A’s trusting B can influence C trusting B or D, and so on); and (iv) how trust can change using generalization reasoning (the fact that it is possible to predict how/when an agent who trusts something/someone will therefore trust something/someone else, before and without a direct experience). These last three aspects are related to \textit{trust influences} in ONTrust. Although it is rather comprehensive, their proposal does not mention the emission of signals to communicate uncertainties nor to indicate trustworthy behavior on the part of the trustee. 

Another related work was proposed by Anantharam ate al. \cite{anantharam2010trust}, who designed a trust ontology to serve as an upper level ontology for use across multiple domains, such as sensor and social networks, e-commerce, collaborative environments, distributed systems, among others. Their trust ontology provides a unique vocabulary for the representation, reasoning and querying over trust relationships existing within a trust network. The authors define the trust relationship as a 6-tuple: trustor, trust type, trust value, trust scope, trust process and trustee. Trust type represents the nature of trust relationship, namely trust in belief and trust in performance, as proposed by Huang and Fox \cite{huang2006ontology}. Trust value quantifies and ranks trust relationships. It is similar to the \textit{trust degree} in ONTrust. Trust scope captures the context for which the trust information is applicable, being analogous to the \textit{trustor’s intention} in ONTrust. Finally, trust process represents the method by which the trust value is created and maintained, such as reputation, provenance, policy and evidence. In general, this latter element is related to the concept of \textit{trustworthiness evidence} in our model. Although they have described the ontology in concrete terms by representing the classes and properties in OWL, the authors acknowledge that their trust ontology is more a taxonomy than a formal semantic specification \cite{thirunarayan2014comparative}.

Lenard et al. \cite{lenard2023exploring} expands upon the ontology proposed by Thirunarayan et al. \cite{thirunarayan2014comparative} to create a trust ontology to be used as a common vocabulary to describe a comparative analysis of Trust Modeling and Management (TMM) approaches in ad-hoc Distributed Wireless Networks (DWNs). Besides representing trust relationships, their ontology defines the concept of trust aggregation, which details how multiple trust values, obtained directly or indirectly from peers, are combined under a unique value. In accordance with Thirunarayan et al. \cite{thirunarayan2014comparative} trust aggregation can be can be reputation-based, policy-based or evidence-based. Additionally, they define a process to compute trust values, which they name trust update, and a process to exchange and publish trust information across the network, which they name trust propagation. Roughly speaking, we can relate this latter to the ONTrust concept of \textit{trust calibration signals}. Similar to Thirunarayan et al. \cite{thirunarayan2014comparative}, they proposal is closer to a taxonomy than a formal semantic specification.

 An important difference between the above-mentioned ontologies and our proposal is that they are not grounded on a foundational ontology. Some of them are built on semantic web languages that give precedence to computational tractability over expressiveness. As discussed by Guizzardi \cite{guizzardi2006role}, a number of semantic interoperability problems cannot be handled by semantic web languages, such as OWL and RDF, as their expressivity is purposefully limited so that they remain computationally efficient.

Table \ref{tab:related-work} summarizes some important aspects of the Reference Ontology of Trust and compare them to the other trust ontologies discussed in this section, such as : (i) the types of trust modeled; (ii) trusted delegation analysis; (iii) quantitative perspective of trust; (iv) the factors that can influence trust; (v) the relation between trust and risk; and (vi) the language used to represent the ontology.

\section{Final Considerations}
\label{sec:conclusions}

This paper presented ONTrust, a reference ontology of trust grounded in the Unified Foundational Ontology \cite{guizzardi2005ontological}, which was developed through an ontological analysis based in a number of results in the literature of computer science, sociology, psychology, cognitive and behavioral sciences. ONTrust formalizes the general concept of trust and characterizes different types of trust, namely social trust, weak trust, strong trust and institution-based trust. ONTrust also provides an ontological account for trusted delegations, for the quantitative perspective of trust and for the factors that can influence trust. Lastly, it leverages the analysis of the behavioral aspect of trust to explain the emergence of risk from trust relations.

In addition, we defined an axiomatization for avoiding unintended model instantiations. This axiomatization was a result of an iterative model simulation approach conducted over ONTrust models. 
This approach consists of transforming OntoUML models into specifications in the logic-based language Alloy \cite{benevides2010validating} and using its analyzer to generate instances of the model automatically. These instantiations were then analyzed to identify the lack of required constraints. The main constraints identified are reflected in the axioms presented in Section \ref{sec:axioms}.

Finally, the ontology was applied to model two case studies from the literature, which exemplified how the distinctions we made contribute to a better understanding, communication and evaluation of trust, both in research and practice.

Our agenda for the future includes the application of our ontology in the solution of practical problems, such as providing sound ontological foundations on trust to support the design of trustworthy systems and for predicting trust (and distrust) in social networks.

Moreover, we plan to delve into trust propagation in ONTrust. As discussed in \cite{huang2006ontology}, the problem of propagating trust is related to the issue of transitivity of trust. As we discussed in Section \ref{sec:ground-trust}, trust is non-transitive, when focusing on the relationship between the trustor and the trustee. It is therefore interesting to overview, assess, and model the conditions of trust propagation within ONTrust. Those conditions shall depend on the types of trust that we have introduced: weak, strong, institution-based trust and trusted delegation.

\begin{landscape}
\begin{table}[]
\tiny
\caption{The Reference Ontology of Trust and related trust ontologies.} 
\label{tab:related-work}
\begin{tabular}{|c|c|c|c|cccc|c|c|}
\hline
\multirow{2}{*}{\textbf{Ontology}}                                                                                  & \multirow{2}{*}{\textbf{Trust Types}}                                                                                     & \multirow{2}{*}{\textbf{\begin{tabular}[c]{@{}c@{}}Trusted \\ Delegation\end{tabular}}} & \multirow{2}{*}{\textbf{\begin{tabular}[c]{@{}c@{}}Quantitative\\ Perspective\end{tabular}}}                            & \multicolumn{4}{c|}{\textbf{Influences}}                                                                                                                                                                                                                                                                                                                                          & \multirow{2}{*}{\textbf{\begin{tabular}[c]{@{}c@{}}Relation to\\ Risk\end{tabular}}}                                             & \multirow{2}{*}{\textbf{\begin{tabular}[c]{@{}c@{}}Modeling\\ Language\end{tabular}}} \\ \cline{5-8}
                                                                                                                    &                                                                                                                           &                                                                                         &                                                                                                                        & \multicolumn{1}{c|}{\textbf{\begin{tabular}[c]{@{}c@{}}Mental\\ Biases\end{tabular}}} & \multicolumn{1}{c|}{\textbf{\begin{tabular}[c]{@{}c@{}}Trustworthiness\\ Evidence\end{tabular}}} & \multicolumn{1}{c|}{\textbf{\begin{tabular}[c]{@{}c@{}}Trust Calibration\\ Signals\end{tabular}}}                 & \textbf{\begin{tabular}[c]{@{}c@{}}Trust\\ Relations\end{tabular}} &                                                                                                                                  &                                                                                       \\ \hline
ONTrust                                                                                                                 & \begin{tabular}[c]{@{}c@{}}Ground Trust\\ Social Trust\\ Institution-based Trust\\ Weak Trust\\ Strong Trust\end{tabular} & YES                                                                                     & \begin{tabular}[c]{@{}c@{}}Trust degree\\ Belief intensity\\ Performance level\\ Manifestation likelihood\end{tabular} & \multicolumn{1}{c|}{YES}                                                              & \multicolumn{1}{c|}{YES}                                                                         & \multicolumn{1}{c|}{\begin{tabular}[c]{@{}c@{}}Trust-warranting \\ signals\\ Uncertainty \\ signals\end{tabular}} & YES                                                                & \begin{tabular}[c]{@{}c@{}}Models the \\ emergence\\ of risk from\\ trust relations\end{tabular}                                 & \begin{tabular}[c]{@{}c@{}}OntoUML\\ \\ gUFO\\ (OWL)\end{tabular}                     \\ \hline
\begin{tabular}[c]{@{}c@{}}FOAF extension \cite{golbeck2003trust}\end{tabular}                   & Trust relation + Intention                                                                                                & NO                                                                                      & \begin{tabular}[c]{@{}c@{}}Trust degree\\ (level of trust)\end{tabular}                                                & \multicolumn{1}{c|}{NO}                                                               & \multicolumn{1}{c|}{NO}                                                                          & \multicolumn{1}{c|}{NO}                                                                                           & YES                                                                & NO                                                                                                                               & RDF/OWL                                                                               \\ \hline
\begin{tabular}[c]{@{}c@{}}PML-T \cite{mcguinness2007pml}\end{tabular}                          & \begin{tabular}[c]{@{}c@{}}Trust relation\\ Belief  (belief relation)\end{tabular}                                        & NO                                                                                      & \begin{tabular}[c]{@{}c@{}}Belief intensity\\ (trust ratings)\end{tabular}                                             & \multicolumn{1}{c|}{NO}                                                               & \multicolumn{1}{c|}{NO}                                                                          & \multicolumn{1}{c|}{NO}                                                                                           & NO                                                                 & NO                                                                                                                               & RDF/OWL                                                                               \\ \hline
\begin{tabular}[c]{@{}c@{}}Trust networks \cite{dokoohaki2008effective}\end{tabular}         & Trust relation + Intention                                                                                                & NO                                                                                      & \begin{tabular}[c]{@{}c@{}}Trust degree\\ (trust leven)\end{tabular}                                                   & \multicolumn{1}{c|}{NO}                                                               & \multicolumn{1}{c|}{YES}                                                                         & \multicolumn{1}{c|}{NO}                                                                                           & NO                                                                 & NO                                                                                                                               & RDF/OWL                                                                               \\ \hline
\begin{tabular}[c]{@{}c@{}}Huang and Fox's\\ Ontology of trust \cite{huang2006ontology}\end{tabular}                  & \begin{tabular}[c]{@{}c@{}}Ground Trust\\ Social Trust\\ Belief (trust in belief)\end{tabular}                            & NO                                                                                      & NO                                                                                                                     & \multicolumn{1}{c|}{NO}                                                               & \multicolumn{1}{c|}{NO}                                                                          & \multicolumn{1}{c|}{NO}                                                                                           & YES                                                                & NO                                                                                                                               & \begin{tabular}[c]{@{}c@{}}Situation\\ calculus\end{tabular}                          \\ \hline
\begin{tabular}[c]{@{}c@{}}Viljanen's\\ Ontology of Trust \cite{viljanen2005towards}\end{tabular}                     & \begin{tabular}[c]{@{}c@{}}Ground Trust\\ Social Trust\end{tabular}                                                       & NO                                                                                      & NO                                                                                                                     & \multicolumn{1}{c|}{NO}                                                               & \multicolumn{1}{c|}{YES}                                                                         & \multicolumn{1}{c|}{NO}                                                                                           & NO                                                                 & \begin{tabular}[c]{@{}c@{}}Relates risk\\ to actions on\\ trusting, but\\ does not detail\\ the emergence\\ of risk\end{tabular} & \begin{tabular}[c]{@{}c@{}}UML\\ OWL\end{tabular}                                  \\ \hline
\begin{tabular}[c]{@{}c@{}}Secure Tropos \cite{giorgini2005modeling}\end{tabular}               & \begin{tabular}[c]{@{}c@{}}Ground Trust\\ Social Trust\\ Strong Trust\end{tabular}                                        & YES                                                                                     & NO                                                                                                                     & \multicolumn{1}{c|}{NO}                                                               & \multicolumn{1}{c|}{NO}                                                                          & \multicolumn{1}{c|}{NO}                                                                                           & NO                                                                 & NO                                                                                                                               & Datalog                                                                               \\ \hline
\begin{tabular}[c]{@{}c@{}}Mechanics of trust \cite{riegelsberger2005mechanics}\end{tabular} & \begin{tabular}[c]{@{}c@{}}Ground Trust\\ Social Trust\\ Institution-based Trust\end{tabular}                             & NO                                                                                      & NO                                                                                                                     & \multicolumn{1}{c|}{NO}                                                               & \multicolumn{1}{c|}{YES}                                                                         & \multicolumn{1}{c|}{\begin{tabular}[c]{@{}c@{}}Trust-warranting\\ signals\end{tabular}}                           & YES                                                                & \begin{tabular}[c]{@{}c@{}}Relates risk\\ to actions on \\ trusting and \\ elaborates on\\ the emergence \\ of risk\end{tabular} & \begin{tabular}[c]{@{}c@{}}Textual\\ descriptions\\ and diagrams\end{tabular}         \\ \hline
\begin{tabular}[c]{@{}c@{}}Castelfranchi and Falcone's\\ Theory of trust\cite{castelfranchi2010trust}\end{tabular} & \begin{tabular}[c]{@{}c@{}}Ground Trust\\ Social Trust\\ Institution-based Trust\end{tabular}                             & YES                                                                                     & YES                                                                                                                    & \multicolumn{1}{c|}{NO}                                                               & \multicolumn{1}{c|}{YES}                                                                         & \multicolumn{1}{c|}{NO}                                                                                           & YES                                                                & \begin{tabular}[c]{@{}c@{}}Emphasizes\\ the relation\\ between trust\\ and risk, but \\ does not \\ model it.\end{tabular}       & \begin{tabular}[c]{@{}c@{}}First-order\\ logic\end{tabular}                           \\ \hline

\begin{tabular}[c]{@{}c@{}}Anantharam et al.\\ Trust model \cite{anantharam2010trust}\end{tabular}                          & \begin{tabular}[c]{@{}c@{}}Ground Trust\\ Social Trust\\ Belief (trust in belief)\end{tabular}                            & NO                                                                                      & \begin{tabular}[c]{@{}c@{}}Trust degree\\ (trust value)\end{tabular}                                                   & \multicolumn{1}{c|}{NO}                                                               & \multicolumn{1}{c|}{YES}                                                                         & \multicolumn{1}{c|}{NO}                                                                                           & YES                                                                & NO                                                                                                                               & \begin{tabular}[c]{@{}c@{}}Graph Diagram\\ OWL\end{tabular}                             \\ \hline

\begin{tabular}[c]{@{}c@{}} Leonard et al.\\ Trust modelling and management \cite{lenard2023exploring}\end{tabular}                        & \begin{tabular}[c]{@{}c@{}}Ground Trust\\ Social Trust\\ Belief (trust in belief)\end{tabular}                            & NO                                                                                      & \begin{tabular}[c]{@{}c@{}}Trust degree\\ (trust value)\end{tabular}                                                   & \multicolumn{1}{c|}{NO}                                                               & \multicolumn{1}{c|}{YES}                                                                         & \multicolumn{1}{c|}{YES}                                                                                          & YES                                                                & NO                                                                                                                               & \begin{tabular}[c]{@{}c@{}}Textual\\ descriptions\\ and diagrams\end{tabular} \\ \hline

\end{tabular}
\end{table}

\end{landscape}

\bibliography{references}

@book{guizzardi2005ontological,
  title={Ontological foundations for structural conceptual models},
  author={Guizzardi, Giancarlo},
  year={2005},
  publisher={Telematica Instituut / CTIT}
}

@article{guizzardi2021ufo,
  title={UFO: Unified Foundational Ontology},
  author={Guizzardi, Giancarlo and Botti Benevides, Alessander and Fonseca, Claudenir M and Porello, Daniele and Almeida, Jo{\~a}o Paulo A and Prince Sales, Tiago},
  journal={Applied Ontology},
  number={Preprint},
  pages={1--44},
  year={2021},
  publisher={IOS Press}
}

@inproceedings{guizzardi2008grounding,
 author={Giancarlo Guizzardi and Ricardo Almeida Falbo and Renata Silva Souza Guizzardi},
 title={Grounding software domain ontologies in the {Unified Foundational Ontology (UFO)}},
 booktitle={11th Ibero-American Conference on Software Engineering (CIbSE)},
 year={2008},
 pages = {127--140},
}

@incollection{guizzardi2010ontology,
  title={Ontology-based transformation framework from {TROPOS} to {AORML}},
  author={Guizzardi, Renata Silva Souza and Guizzardi, Giancarlo},
  booktitle={Social modeling for requirements engineering},
  pages={547--570},
  year={2010},
  publisher={The MIT Press}
}

@inproceedings{guizzardi2013towards,
  title={Towards ontological foundations for the conceptual modeling of events},
  author={Giancarlo Guizzardi and Wagner, Gerd and Ricardo Almeida Falbo and Renata Silva Souza Guizzardi and Almeida, João Paulo Andrade},
  booktitle={32nd International Conference on Conceptual Modeling {(ER)}},
  pages={327--341},
  year={2013},
  organization={Springer},
}

@article{guizzardi2015towards,
  title={Towards ontological foundations for conceptual modeling: the {Unified Foundational Ontology (UFO)} story},
  author={Giancarlo Guizzardi and Wagner, Gerd and Almeida, João Paulo Andrade and Renata Silva Souza Guizzardi},
  journal={Applied ontology},
  volume={10},
  number={3-4},
  pages={259--271},
  year={2015},
  publisher={IOS Press}
}

@article{azevedo2015modeling,
  title={Modeling resources and capabilities in enterprise architecture: A well-founded ontology-based proposal for {ArchiMate}},
  author={{Azevedo, C.L.B. et al.}},
  journal={Information systems},
  volume={54},
  pages={235--262},
  year={2015},
  publisher={Elsevier},
}

@InCollection{varzi2007omissions,
  title={Omissions and causal explanations},
  author={Varzi, Achille C},
  editors={F. Castellani and J. Quitterer},
  booktitle={Agency and Causation in the Human Sciences}, 
  publisher = {Paderborn, Germany: Mentis Verlag},
  pages ={155-67},
  year={2007}
}

@inproceedings{guizzardi2006role,
  title={The role of foundational ontologies for conceptual modeling and domain ontology representation},
  author={Guizzardi, Giancarlo},
  booktitle={2006 7th International Baltic conference on databases and information systems},
  pages={17--25},
  year={2006},
  organization={IEEE}
}

@incollection{pribbenow2002meronymic,
  title={Meronymic relationships: From classical mereology to complex part-whole relations},
  author={Pribbenow, Simone},
  booktitle={The semantics of relationships},
  pages={35--50},
  year={2002},
  publisher={Springer}
}

@article{gerstl1995midwinters,
  title={Midwinters, end games, and body parts: a classification of part-whole relations},
  author={Gerstl, Peter and Pribbenow, Simone},
  journal={International journal of human-computer studies},
  volume={43},
  number={5-6},
  pages={865--889},
  year={1995},
  publisher={Elsevier}
}

@inproceedings{sales2017fleet,
  title={``{Is} It a Fleet or a Collection of Ships?'': Ontological Anti-patterns in the Modeling of Part-Whole Relations},
  author={Sales, Tiago Prince and Guizzardi, Giancarlo},
  booktitle={European Conference on Advances in Databases and Information Systems},
  pages={28--41},
  year={2017},
  organization={Springer}
}

@techreport{gufo2020,
  title={{gUFO: A Lightweight Implementation of the Unified Foundational Ontology (UFO)}},
  author={Almeida, J.P.A and Guizzardi, Giancarlo and Sales, Tiago Prince and Falbo, R.A.},
  year={2020},
  institution={{ Ontology \& Conceptual Modeling Research Group (NEMO) - Federal University of Espirito Santo }}
}

@inproceedings{guizzardi2011design,
  title={{Design patterns and inductive modeling rules to support the construction of ontologically well-founded conceptual models in OntoUML}},
  author={Guizzardi, Giancarlo and das Gra{\c{c}}as, Alex Pinheiro and Guizzardi, Renata SS},
  booktitle={CAISE Workshops},
  pages={402--413},
  year={2011},
  organization={Springer}
}

@article{sales2015ontological,
  title={{Ontological anti-patterns: Empirically uncovered error-prone structures in ontology-driven conceptual models}},
  author={Sales, Tiago Prince and Guizzardi, Giancarlo},
  journal={Data \& Knowledge Eng.},
  volume={99},
  pages={72--104},
  year={2015},
  publisher={Elsevier}
}

@article{benevides2010validating,
  title={{Validating Modal Aspects of OntoUML Conceptual Models Using Automatically Generated Visual World Structures}},
  author={Benevides, Alessander Botti and Guizzardi, Giancarlo and Braga, Bernardo Ferreira Bastos and Almeida, Joao Paulo A},
  journal={J. Univers. Comput. Sci.},
  volume={16},
  number={20},
  pages={2904--2933},
  year={2010}
}

@article{barcelos2013automated,
  title={{An Automated Transformation from OntoUML to OWL and SWRL}},
  author={Barcelos, Pedro Paulo F and dos Santos, Victor Amorim and Silva, Freddy Brasileiro and Monteiro, Maxwell E and Garcia, Anilton Salles},
  journal={Ontobras},
  volume={1041},
  pages={130--141},
  year={2013}
}

@inproceedings{rybola2016towards,
  title={{Towards OntoUML for software engineering: transformation of anti-rigid sortal types into relational databases}},
  author={Rybola, Zden{\v{e}}k and Pergl, Robert},
  booktitle={International Conference on Model and Data Engineering},
  pages={1--15},
  year={2016},
  organization={Springer}
}

@inproceedings{fumagalli2020towards,
  title={Towards automated support for conceptual model diagnosis and repair},
  author={Fumagalli, Mattia and Sales, Tiago Prince and Guizzardi, Giancarlo},
  booktitle={International Conference on Conceptual Modeling},
  pages={15--25},
  year={2020},
  organization={Springer}
}

@article{moltmann202019,
  title={Variable Objects and Truthmaking Friederike Moltmann},
  author={Moltmann, Friederike},
  journal={Metaphysics, Meaning, and Modality},
  pages={368--394},
  year={2020},
  publisher={Oxford University Press}
}

@book{castelfranchi2010trust,
  title={Trust theory: A socio-cognitive and computational model},
  author={Castelfranchi, Christiano and Falcone, Rino},
  volume={18},
  year={2010},
  publisher={John Wiley \& Sons}
}

@inproceedings{falcone2004trust,
  title={Trust dynamics: How trust is influenced by direct experiences and by trust itself},
  author={Falcone, Rino and Castelfranchi, Cristiano},
  booktitle={Proceedings of the Third International Joint Conference on Autonomous Agents and Multiagent Systems, 2004. AAMAS 2004.},
  pages={740--747},
  year={2004},
  organization={IEEE}
}

@book{barber1983logic,
  title={The logic and limits of trust},
  author={Barber, Bernard},
  year={1983},
  publisher={Rutgers University Press},
  edition={1}
}

@book{luhmann2018trust,
  title={Trust and power},
  author={Luhmann, Niklas},
  year={2018},
  publisher={John Wiley \& Sons}
}

@article{williamson1993calculativeness,
  title={Calculativeness, trust, and economic organization},
  author={Williamson, Oliver E},
  journal={The journal of law and economics},
  volume={36},
  number={1, Part 2},
  pages={453--486},
  year={1993},
  publisher={The University of Chicago Press}
}

@article{rotter1967new,
  title={A new scale for the measurement of interpersonal trust},
  author={Rotter, Julian B},
  journal={Journal of personality},
  volume={35},
  number={4},
  pages={651--665},
  year={1967},
  publisher={Wiley Online Library}
}

@book{tyler2006people,
  title={Why people obey the law},
  author={Tyler, Tom R},
  year={2006},
  publisher={Princeton University Press}
}

@article{cross2004law,
  title={Law and trust},
  author={Cross, Frank B},
  journal={Georgetown Law Journal},
  volume={93},
  pages={1457},
  year={2005},
  publisher={HeinOnline}
}

@inproceedings{moyano2012conceptual,
  title={A conceptual framework for trust models},
  author={Moyano, Francisco and Fernandez-Gago, Carmen and Lopez, Javier},
  booktitle={International Conference on Trust, Privacy and Security in Digital Business},
  pages={93--104},
  year={2012},
  organization={Springer}
}

@article{gambetta2000can,
  title={Can we trust trust},
  author={Gambetta, Diego and others},
  journal={Trust: Making and breaking cooperative relations},
  volume={13},
  pages={213--237},
  year={2000}
}

@article{mayer1995integrative,
  title={An integrative model of organizational trust},
  author={Mayer, Roger C and Davis, James H and Schoorman, F David},
  journal={Academy of management review},
  volume={20},
  number={3},
  pages={709--734},
  year={1995},
  publisher={Academy of Management Briarcliff Manor, NY 10510}
}

@article{rousseau1998not,
  title={Not so different after all: A cross-discipline view of trust},
  author={Rousseau, Denise M and Sitkin, Sim B and Burt, Ronald S and Camerer, Colin},
  journal={Academy of management review},
  volume={23},
  number={3},
  year={1998},
  publisher={Academy of Management Briarcliff Manor, NY 10510}
}

@incollection{mcknight2001trust,
  title={Trust and distrust definitions: One bite at a time},
  author={McKnight, D Harrison and Chervany, Norman L},
  booktitle={Trust in Cyber-societies},
  pages={27--54},
  year={2001},
  publisher={Springer}
}

@inproceedings{mcknight2012events,
  title={How events affect trust: A baseline information processing model with three extensions},
  author={McKnight, D Harrison and Liu, Peng and Pentland, Brian T},
  booktitle={IFIP International Conference on Trust Management},
  pages={217--224},
  year={2012},
  organization={Springer}
}

@article{lewis1985trust,
  title={Trust as a social reality},
  author={Lewis, J David and Weigert, Andrew},
  journal={Social forces},
  volume={63},
  number={4},
  year={1985},
  publisher={Oxford University Press}
}

@inproceedings{giorgini2005modeling,
  title={Modeling social and individual trust in requirements engineering methodologies},
  author={Giorgini, Paolo and Massacci, Fabio and Mylopoulos, John and Zannone, Nicola},
  booktitle={International Conference on Trust Management},
  pages={161--176},
  year={2005},
  organization={Springer}
}

@article{bresciani2004tropos,
  title={Tropos: An agent-oriented software development methodology},
  author={Bresciani, Paolo and Perini, Anna and Giorgini, Paolo and Giunchiglia, Fausto and Mylopoulos, John},
  journal={Autonomous Agents and Multi-Agent Systems},
  volume={8},
  number={3},
  pages={203--236},
  year={2004},
  publisher={Springer}
}

@inproceedings{viljanen2005towards,
  title={Towards an ontology of trust},
  author={Viljanen, Lea},
  booktitle={International Conference on Trust, Privacy and Security in Digital Business},
  pages={175--184},
  year={2005},
  organization={Springer}
}

@inproceedings{huang2006ontology,
  title={An ontology of trust: formal semantics and transitivity},
  author={Huang, Jingwei and Fox, Mark S},
  booktitle={Proceedings of the 8th international conference on Electronic commerce: The new e-commerce: innovations for conquering current barriers, obstacles and limitations to conducting successful business on the internet},
  pages={259--270},
  year={2006},
  organization={ACM}
}

@article{dokoohaki2008effective,
  title={Effective design of trust ontologies for improvement in the structure of socio-semantic trust networks},
  author={Dokoohaki, Nima and Matskin, Mihhail},
  journal={International Journal On Advances in Intelligent Systems},
  volume={1},
  number={1942-2679},
  pages={23--42},
  year={2008},
  publisher={Citeseer}
}

@inproceedings{golbeck2003trust,
  title={Trust networks on the semantic web},
  author={Golbeck, Jennifer and Parsia, Bijan and Hendler, James},
  booktitle={International workshop on cooperative information agents},
  year={2003},
  organization={Springer}
}

@inproceedings{mcguinness2007pml,
  title={PML 2: A Modular Explanation Interlingua.},
  author={McGuinness, Deborah L and Ding, Li and Da Silva, Paulo Pinheiro and Chang, Cynthia},
  booktitle={ExaCt},
  pages={49--55},
  year={2007}
}

@incollection{jacquette2013belief,
  title={Belief state intensity},
  author={Jacquette, Dale},
  booktitle={New Essays on Belief},
  pages={209--229},
  year={2013},
  publisher={Springer}
}

@article{marsh1994formalising,
  title={{Formalising trust as a computational concept. Ph.D. thesis}},
  author={Marsh, Stephen Paul},
  journal={University of Stirling, Department of Computer Science and Mathematics},
  year={1994},
  publisher={University of Stirling}
}

@article{ennew2007measuring,
  title={Measuring trust in financial services: The trust index},
  author={Ennew, Christine and Sekhon, Harjit},
  journal={Consumer Policy Review},
  volume={17},
  number={2},
  pages={62},
  year={2007},
  publisher={CONSUMERS'ASSOCIATION}
}

@inproceedings{agudo2008model,
author="Agudo, Isaac
and Fernandez-Gago, Carmen
and Lopez, Javier",
editor="Furnell, Steven
and Katsikas, Sokratis K.
and Lioy, Antonio",
title="A Model for Trust Metrics Analysis",
booktitle="Trust, Privacy and Security in Digital Business",
year="2008",
publisher="Springer",
pages="28--37",
isbn="978-3-540-85735-8"
}

@article{riegelsberger2005mechanics,
  title={The mechanics of trust: A framework for research and design},
  author={Riegelsberger, Jens and Sasse, M Angela and McCarthy, John D},
  journal={International Journal of Human-Computer Studies},
  volume={62},
  number={3},
  pages={381--422},
  year={2005},
  publisher={Elsevier}
}

@misc{cnn2020binden,
    title={Biden receives first dose of Covid-19 vaccine on live television},
    author = {Kate Sullivan},
    year={2020},
    url = {https://edition.cnn.com/2020/12/21/politics/bidens-coronavirus-vaccination/index.html},
    note = {{Accessed: 2021-01-10}}
}

@article{tomsett2020rapid,
  title={{Rapid trust calibration through interpretable and uncertainty-aware AI}},
  author={{Tomsett, Richard et al.}},
  journal={Patterns},
  volume={1},
  number={4},
  year={2020},
  publisher={Elsevier}
}

@article{batteux2021negative,
  title={{The negative consequences of failing to communicate uncertainties during a pandemic: The case of COVID-19 vaccines}},
  author={Batteux, Eleonore and Avri, Bilovich and Johnson, Samuel GB and Tuckett, David},
  journal={medRxiv},
  year={2021},
  publisher={Cold Spring Harbor Laboratory Press}
}

@article{fischer2013individuals,
  title={{Why do individuals respond to fraudulent scam communications and lose money? The psychological determinants of scam compliance}},
  author={Fischer, Peter and Lea, Stephen EG and Evans, Kath M},
  journal={Journal of Applied Social Psychology},
  volume={43},
  number={10},
  pages={2060--2072},
  year={2013},
  publisher={Wiley Online Library}
}

@article{chen2012enhance,
  title={Enhance green purchase intentions: The roles of green perceived value, green perceived risk, and green trust},
  author={Chen, Yu-Shan and Chang, Ching-Hsun},
  journal={Management Decision},
  year={2012},
  publisher={Emerald Group Publishing Limited}
}

@article{hleg2019ethics,
  title={{Ethics Guidelines for Trustworthy AI}},
author={{Hleg, A.I.}},
  journal={B-1049 Brussels},
  year={2019}
}

@article{dunn2005feeling,
  title={Feeling and believing: the influence of emotion on trust.},
  author={Dunn, Jennifer R and Schweitzer, Maurice E},
  journal={Journal of personality and social psychology},
  volume={88},
  number={5},
  pages={736},
  year={2005},
  publisher={American Psychological Association}
}

@inproceedings{sales2018cover,
  author = {Tiago Prince Sales and Fernanda Bai{\~a}o and Giancarlo Guizzardi and Nicola Guarino and John Mylopoulos},
  title = {The Common Ontology of Value and Risk},
  booktitle={{37th International Conference on Conceptual Modeling (ER)}},
  year = {2018},
  pages={121--135},
  volume={11157},
  organization={Springer}
}

@inproceedings{amaral2019towards,
  title={{Towards a Reference Ontology of Trust}},
  author={Amaral, Glenda and Sales, Tiago Prince and Guizzardi, Giancarlo and Porello, Daniele},
  booktitle={{On the Move to Meaningful Internet Systems: OTM 2019 Conferences}},
  pages={3--21},
  year={2019},
  organization={Springer}
}

@inproceedings{amaral2021ROT,
  title={{Ontological Foundations for Trust Management: Extending the Reference Ontology of Trust}},
  author={Amaral, Glenda and Sales, Tiago Prince and Guizzardi, Giancarlo},
  booktitle = {15th International Workshop on Value Modelling and Business Ontologies (VMBO)},
  year={2021},
  pages={12--22},
  volume={2835},
  organization={{CEUR-WS.org}}
}

@inproceedings{amaral2020TPL,
  title={{Modeling Trust in Enterprise Architecture: A Pattern Language for ArchiMate}},
  author={Amaral, Glenda and Sales, Tiago Prince and Guizzardi, Giancarlo and Almeida, Joao Paulo A and Porello, Daniele},
  booktitle={The Practice of Enterprise Modeling (PoEM)},
  pages={73--89},
  year={2020},
  organization={Springer}
}

@inproceedings{amaral2020trustworthiness,
  title={{Ontology-based Modeling and Analysis of Trustworthiness Requirements: Preliminary Results}},
  author={Amaral, Glenda and Guizzardi, Renata and Guizzardi, Giancarlo and Mylopoulos, John},
  booktitle={39th International Conference on Conceptual Modeling (ER)},
  pages={342--352},
  year={2020},
  organization={Springer}
}

@inproceedings{amaral2021trustworthinessPix,
  title={{Trustworthiness Requirements: The Pix Case Study}},
  author={Amaral, Glenda and Guizzardi, Renata and Guizzardi, Giancarlo and Mylopoulos, John},
  booktitle={40th International Conference on Conceptual Modeling (ER)},
  pages={257–-267},
  year={2021},
  organization={Springer}
}

@inproceedings{amaral2022trustCBDC,
  title={{Ontological Foundations for Trust Dynamics: The Case of Central Bank Digital Currency Ecosystems}},
  author={Amaral, Glenda and Sales, Tiago Prince and Guizzardi, Giancarlo},
  booktitle={16th International Conference on Research Challenges in Information Science (RCIS)},
  year={2022},
  organization={Springer}
}

@inproceedings{baratella2023many,
  title={The Many Facets of Trust},
 author={Baratella, Riccardo and Amaral, Glenda and Sales, Tiago Prince and Guizzardi, Renata and Guizzardi, Giancarlo},
  booktitle={FOIS},
  pages={},
  year={2023}
}

@article{avgerou2013explaining,
  title={Explaining trust in IT-mediated elections: A case study of e-voting in Brazil},
  author={Avgerou, Chrisanthi},
  journal={Journal of the Association for Information Systems},
  volume={14},
  number={8},
  pages={2},
  year={2013}
}

@article{juravle2020trust,
  title={Trust in artificial intelligence for medical diagnoses},
  author={Juravle, Georgiana and Boudouraki, Andriana and Terziyska, Miglena and Rezlescu, Constantin},
  journal={Progress in brain research},
  volume={253},
  pages={263--282},
  year={2020},
  publisher={Elsevier}
}

@article{rosen2013media,
  title={The media and technology usage and attitudes scale: An empirical investigation},
  author={Rosen, Larry D and Whaling, Kelly and Carrier, L Mark and Cheever, Nancy A and Rokkum, Jeffrey},
  journal={Computers in human behavior},
  volume={29},
  number={6},
  pages={2501--2511},
  year={2013},
  publisher={Elsevier}
}

@inproceedings{anantharam2010trust,
  title={Trust model for semantic sensor and social networks: A preliminary report},
  author={Anantharam, Pramod and Henson, Cory A and Thirunarayan, Krishnaprasad and Sheth, Amit P},
  booktitle={Proceedings of the IEEE 2010 National Aerospace \& Electronics Conference},
  pages={1--5},
  year={2010},
  organization={IEEE}
}

@article{thirunarayan2014comparative,
  title={Comparative trust management with applications: Bayesian approaches emphasis},
  author={Thirunarayan, Krishnaprasad and Anantharam, Pramod and Henson, Cory and Sheth, Amit},
  journal={Future Generation Computer Systems},
  volume={31},
  pages={182--199},
  year={2014},
  publisher={Elsevier}
}

@article{lenard2023exploring,
  title={Exploring Trust Modelling and Management Techniques in the Context of Distributed Wireless Networks: A Literature Review},
  author={Lenard, Teri and Collen, Anastasija and Benyahya, Meriem and Nijdam, Niels Alexander and Genge, B{\'e}la},
  journal={IEEE Access},
  year={2023},
  publisher={IEEE}
}

@Online{Melnick2024,
 author = {Melnick, Kyle},
 year = {2024},
 title = {Air Canada chatbot promised a discount. Now the airline has to pay it.},
 journal = {The Washington Post},
 url = {https://www.washingtonpost.com/travel/2024/02/18/air-canada-airline-chatbot-ruling/},
 urldate = {2024-02-18}
}

@article{schar2021decentralized,
  title={Decentralized finance: On blockchain and smart contract-based financial markets},
  author={Sch{\"a}r, Fabian},
  journal={FRB of St. Louis Review},
  volume={103},
  year={2021}
}

@article{zetzsche2020decentralized,
  title={{Decentralized finance (DeFi)}},
  author={Zetzsche, Dirk A and Arner, Douglas W and Buckley, Ross P},
  journal={IIEL Issue Brief},
  volume={2},
  year={2020}
}

@article{omg2003ocl,
  title={{Unified Modelling Language: Object Constraint Language version 2.0, ptc/03-10-04}},
  author={{Object Managment Group}},
  journal={},
  volume={},
  year={2003}
}

@book{searle1995construction,
  title={The Construction of Social Reality},
  author={Searle, J.R.},
  year={1995},
  publisher={Free Press}
}

@article{morales2015ontology,
  title={An ontology of online user feedback in software engineering},
  author={Morales-Ramirez, Itzel and Perini, Anna and Guizzardi, Renata SS},
  journal={Applied Ontology},
  volume={10},
  number={3-4},
  pages={297--330},
  year={2015},
  publisher={IOS Press}
}

@book{simpson2023trust,
  title={Trust: A philosophical study},
  author={Simpson, Thomas W},
  year={2023},
  publisher={Oxford University Press}
}

@book{simon2020routledge,
  title={The Routledge handbook of trust and philosophy},
  author={Simon, Judith},
  year={2020},
  publisher={Routledge}
}

@article{mujdricza2019roots,
  title={The roots of trust},
  author={M{\'u}jdricza, Ferenc},
  journal={European Journal of Mental Health},
  volume={14},
  number={1},
  pages={109--142},
  year={2019},
  publisher={Semmelweis Egyetem Ment{\'a}lhigi{\'e}n{\'e} Int{\'e}zet}
}

@article{pytlikzillig2016consensus,
  title={Consensus on conceptualizations and definitions of trust: Are we there yet?},
  author={PytlikZillig, Lisa M and Kimbrough, Christopher D},
  journal={Interdisciplinary perspectives on trust: Towards theoretical and methodological integration},
  pages={17--47},
  year={2016},
  publisher={Springer}
}

@article{kelsall2024towards,
  title={Towards a non-reliance commitment account of trust},
  author={Kelsall, Joshua},
  journal={The Journal of Value Inquiry},
  pages={1--17},
  year={2024},
  publisher={Springer}
}

@article{afroogh2023probabilistic,
  title={A probabilistic theory of trust concerning artificial intelligence: can intelligent robots trust humans?},
  author={Afroogh, Saleh},
  journal={AI and Ethics},
  volume={3},
  number={2},
  pages={469--484},
  year={2023},
  publisher={Springer}
}

@article{gille2020we,
  title={What we talk about when we talk about trust: theory of trust for AI in healthcare},
  author={Gille, Felix and Jobin, Anna and Ienca, Marcello},
  journal={Intelligence-Based Medicine},
  volume={1},
  pages={100001},
  year={2020},
  publisher={Elsevier}
}

@article{lukyanenko2022trust,
  title={Trust in artificial intelligence: From a Foundational Trust Framework to emerging research opportunities},
  author={Lukyanenko, Roman and Maass, Wolfgang and Storey, Veda C},
  journal={Electronic Markets},
  volume={32},
  number={4},
  pages={1993--2020},
  year={2022},
  publisher={Springer}
}

@article{glikson2020human,
  title={Human trust in artificial intelligence: Review of empirical research},
  author={Glikson, Ella and Woolley, Anita Williams},
  journal={Academy of management annals},
  volume={14},
  number={2},
  pages={627--660},
  year={2020},
  publisher={Briarcliff Manor, NY}
}

@book{hidalgo2021humans,
  title={How humans judge machines},
  author={Hidalgo, C{\'e}sar A and Orghian, Diana and Canals, Jordi Albo and De Almeida, Filipa and Martin, Natalia},
  year={2021},
  publisher={MIT Press}
}

@article{yang2022user,
  title={User trust in artificial intelligence: A comprehensive conceptual framework},
  author={Yang, Rongbin and Wibowo, Santoso},
  journal={Electronic Markets},
  volume={32},
  number={4},
  pages={2053--2077},
  year={2022},
  publisher={Springer}
}

@article{eyal2024trust,
  title={Trust is a Verb: A Crtical Reconstruction of the Sociological Theory of Trust},
  author={Eyal, Gil and Capotescu, Cristian and Au, Larry},
  journal={International Journal for Sociological Debate},
  volume={18},
  number={2},
  year={2024}
}

@article{ur2019trust,
  title={Trust in blockchain cryptocurrency ecosystem},
  author={ur Rehman, Muhammad Habib and Salah, Khaled and Damiani, Ernesto and Svetinovic, Davor},
  journal={IEEE Transactions on Engineering Management},
  volume={67},
  number={4},
  pages={1196--1212},
  year={2019},
  publisher={IEEE}
}

@inproceedings{oliveira2025ontological,
  title={An Ontological Model of the Phishing Attack Process},
  author={Oliveira, {\'I}talo and Wagner, Gerd and Amaral, Glenda and Sales, Tiago Prince and Bull{\'e}e, Jan-Willem and Junger, Marianne and Sarmah, Dipti K and Daneva, Maya and Guizzardi, Giancarlo},
  booktitle={International Conference on Business Process Modeling, Development and Support},
  pages={274--289},
  year={2025},
  organization={Springer}
}

@article{amaral2025combining,
  title={Combining Neural Empathy-Aware Behavior Trees with Knowledge Graphs for Affective Human-AI Teaming},
  author={Amaral, Glenda and Costantini, Stefania and De Gasperis, Giovanni and De Lauretis, Lorenzo and Dell'Acqua, Pierangelo and Guizzardi, Giancarlo and Gullo, Francesco and Rafanelli, Andrea},
  journal={IEEE Transactions on Affective Computing},
  year={2025},
  publisher={IEEE}
}

@inproceedings{amaral2025unpacking,
  title={Unpacking Trust: An Ontological Framework for Information Trustworthiness in Decision-Making},
  author={Amaral, Glenda and Santos, Veronica dos and Haeusler, Edward Hermann and Guizzardi, Giancarlo and Schwabe, Daniel and Lifschitz, Sergio},
  booktitle={International Conference on Conceptual Modeling},
  pages={181--190},
  year={2025},
  organization={Springer}
}

@article{touameur2025guitares,
  title={GUITARES: graph attention network for building knowledge graph-based trust-aware recommender systems},
  author={Touameur, Ouissem and Harrag, Fouzi and Bellatreche, Ladjel},
  journal={The Journal of Supercomputing},
  volume={81},
  number={11},
  pages={1199},
  year={2025},
  publisher={Springer}
}

@inproceedings{ding2024conceptual,
  title={A Conceptual Model for Blockchain-Based Trust in Digital Ecosystems (Short Paper)},
  author={Ding, Yuntian and Herbaut, Nicolas and Negru, Daniel},
  booktitle={International Conference on Advanced Information Systems Engineering},
  pages={18--24},
  year={2024},
  organization={Springer}
}

@inproceedings{shishkov2023incorporating,
  title={Incorporating trust into context-aware services},
  author={Shishkov, Boris and Fill, Hans-Georg and Ivanova, Krassimira and van Sinderen, Marten and Verbraeck, Alexander},
  booktitle={International Symposium on Business Modeling and Software Design},
  pages={92--109},
  year={2023},
  organization={Springer}
}

@inproceedings{lanasri2020trust,
  title={Trust-aware curation of linked open data logs},
  author={Lanasri, Dihia and Khouri, Selma and Bellatreche, Ladjel},
  booktitle={International Conference on Conceptual Modeling},
  pages={604--614},
  year={2020},
  organization={Springer}
}

@article{regona2026building,
  title={Building Trust in Artificial Intelligence: A Systematic Review through the Lens of Trust Theory},
  author={Regona, Massimo and Yigitcanlar, Tan and Hon, Carol and Teo, Melissa},
  journal={ACM Computing Surveys},
  year={2026},
  publisher={ACM New York, NY}
}

\appendix

\begin{landscape}

\subsection*{Appendix 1 - Tables from the ``Trust in IT-mediated elections'' case}

\footnotesize
\begin{longtable}{|p{0,2cm}|p{2cm}|p{5cm}|p{8cm}|}
\caption{Brazilian electors trust in e-voting.} 
\label{tab:trust-types-table}
\endfirsthead
\endhead
\hline
\textbf{\#} & \textbf{Concepts} & \textbf{Instantiation(s) in the example} & \textbf{Relationship(s)} 
\\ \hline

1 & Trustor & 
\begin{itemize}[leftmargin=*]
\item[] electors 
\end{itemize} & 
\begin{itemize}[leftmargin=*]
\item[] \textbf{trust} \textsf{e-voting} ecosystem for choosing their leaders 
\item[] \textbf{intend} to choose their leaders
\item[] \textbf{believe} that the \textsf{e-voting} is conducted lawfully and competently
\item[] \textbf{believe} that \textsf{e-voting} results are an accurate representation of citizens’ preferences
\item[] \textbf{believe} that \textsf{e-voting} ecosystem is secure, privacy preserving and accurate
\item[] \textbf{trust} the \textsf{Electoral Justice System} to support e-voting
\end{itemize} \\ \hline

2 & Trustee &
\begin{itemize}[leftmargin=*]
\item[] e-voting ecosystem, composed by the e-voting technology and the  electoral authorities
\item[] Electoral Justice System
\end{itemize} &
\begin{itemize}[leftmargin=*]
\item[] \textbf{is trusted} by \textsf{electors} 
\end{itemize} \\ \hline

3 & Ground Trust &
\begin{itemize}[leftmargin=*]
\item[] complex mental state of \textsf{electors} regarding the \textsf{e-voting} ecosystem and its behavior, w.r.t. their intention of choosing their leaders.
\end{itemize}&
\begin{itemize}[leftmargin=*]
\item[] \textbf{is composed by} the \textsf{electors}’ \textsf{intention} of choosing their leaders and the set of \textsf{beliefs} about the \textsf{e-voting ecosystem} \textsf{capabilities} and \textsf{vulnerabilities} (\#5 and \#7, respectively) 
\end{itemize} \\ \hline

4 & Intention &
\begin{itemize}[leftmargin=*]
\item[] choose their leaders 
\end{itemize} &
\begin{itemize}[leftmargin=*]
\item[] \textbf{inheres in} the \textsf{electors} 
\item[] \textbf{is component} part of \textsf{(ground) trust} of \textsf{electors} in the \textsf{e-voting ecosystem}
\end{itemize} \\ \hline

5 & Capability Belief &
\begin{itemize}[leftmargin=*]
\item[] e-voting is conducted lawfully
\item[] e-voting is conducted competently
\item[] e-voting results are an accurate representation of citizens’ preferences
\item[] e-voting ecosystem is secure
\item[] e-voting ecosystem preserves elector’s privacy
\item[] e-voting ecosystem is accurate
\end{itemize}
&
\begin{itemize}[leftmargin=*]
\item[] \textbf{is component part} of \textsf{(ground) trust} of \textsf{electors} in the \textsf{e-voting ecosystem}
\end{itemize} \\ \hline

6 & Moment type (capability type) &
\begin{itemize}[leftmargin=*]
\item[] competence to properly conduct the electoral process
\item[] compliance with laws
\item[] security mechanisms
\item[] privacy preserving mechanisms
\item[] functionalities to provide accurate results
\end{itemize}
&
\textbf{inheres} in the \textsf{e-voting ecosystem} (substantial type) \\ \hline

7 & Vulnerability Belief &
\begin{itemize}[leftmargin=*]
\item[] the e-voting ecosystem has mechanisms to prevent security breaches
\item[] the e-voting ecosystem has mechanisms to prevent data leaks
\item[] the e-voting ecosystem has mechanisms to avoid technology malfunctions
\item[] the e-voting ecosystem has mechanisms to avoid technology inefficiencies
\item[] the e-voting ecosystem has mechanisms to prevent frauds
\end{itemize}
&
\begin{itemize}[leftmargin=*]
\item[] \textbf{is component part} of \textsf{(ground) trust} of \textsf{electors} in the \textsf{e-voting ecosystem}
\end{itemize} \\ \hline

8 & Moment type (vulnerability type) &
\begin{itemize}[leftmargin=*]
\item[] security breaches
\item[] technology inefficiencies
\item[] technology malfunctioning
\item[] technology fraud
\end{itemize}
&
\begin{itemize}[leftmargin=*]
\item[] \textbf{inheres} in the \textsf{e-voting ecosystem} (substantial type)
\end{itemize} \\ \hline 

9 & Social Trust &
\begin{itemize}[leftmargin=*]
\item[] \textsf{electors}’ \textsf{(social) trust} in \textsf{electoral authorities} (a component of the e-voting ecosystem) to choose their leaders
\end{itemize} &
\begin{itemize}[leftmargin=*]
\item[] specialization of \textsf{ground trust} (\#3) that includes the \textsf{electors}’ \textsf{belief} that the \textsf{electoral authorities} intend to conduct the e-voting lawfully
\end{itemize} \\ \hline  

10 & Intention Belief &
\begin{itemize}[leftmargin=*]
\item[] electoral authorities intend to conduct the e-voting according to the law
\item[] electoral authorities have no intention to abuse their power
\end{itemize}
&
\begin{itemize}[leftmargin=*]
\item[] \textbf{is component part} of \textsf{(social) trust} of \textsf{electors} in the \textsf{e-voting ecosystem}
\end{itemize} \\ \hline

11 & Moment type (intention type) &
\begin{itemize}[leftmargin=*]
\item[] conduct the e-voting according to the law
\end{itemize}
&
\begin{itemize}[leftmargin=*]
\item[] \textbf{inheres} in the \textsf{electoral authorities} (agent type)
\end{itemize} \\ \hline

12 & Strong Trust &
\begin{itemize}[leftmargin=*]
\item[] \textsf{electors}’ \textsf{(strong) trust} in the \textsf{e-voting ecosystem} to choose their leaders
\end{itemize}
&
\begin{itemize}[leftmargin=*]
\item[] trust that (i) includes the \textsf{electors}’ \textsf{belief} that the electoral authorities are committed to conduct the e-voting according to the law, as well as the \textsf{e-voting ecosystem} is committed to providing accurate results; and (ii) is grounded on a public agreement in which the \textsf{electoral authorities} are committed to conduct the e-voting according to the law. 
\end{itemize} \\ \hline

13 & Social Commitment Belief &
\begin{itemize}[leftmargin=*]
\item[] electoral authorities are committed to to conducting the e-voting according to the law
\item[] e-voting ecosystem is committed to providing accurate results
\end{itemize}
&
\begin{itemize}[leftmargin=*]
\item[] \textbf{is component part} of \textsf{(social) trust} of \textsf{electors} in the \textsf{e-voting ecosystem}
\end{itemize} \\ \hline

14 & Trustee Commitment  &
\begin{itemize}[leftmargin=*]
\item[] electoral authorities are committed to conducting the e-voting according to the law
\end{itemize}
&
\begin{itemize}[leftmargin=*]
\item[] \textbf{is component part} of a public \textsf{agreement} in which the \textsf{electoral authorities} commit to conduct the e-voting according to the law
\end{itemize} \\ \hline

15 & Institution-based Trust &
\begin{itemize}[leftmargin=*]
\item[] \textsf{electors}’ \textsf{trust} in the Electoral Justice System to support e-voting
\end{itemize} 
&
\begin{itemize}[leftmargin=*]
\item[] \textbf{is composed by} the \textsf{electors}' \textsf{intention} of choosing their leaders and a set of \textsf{beliefs} about the Electoral Justice System (for example, the existence of laws, processes, regulatory mechanisms and other protective structures that are conducive to situational success). 
\end{itemize} \\ \hline

\end{longtable}

\vspace{25pt}

\begin{longtable}{|p{0,2cm}|p{2cm}|p{13cm}|p{2cm}|}
\caption{Factors that influence trust in e-voting.} 
\label{tab:trust-influences-table}
\endfirsthead
\endhead
\hline
\textbf{\#} & \textbf{Concepts} & \textbf{Instantiation(s) in the example} & \textbf{Relationship(s)} 
\\ \hline

16 & Trust Relation & 
\begin{itemize}[leftmargin=*]
\item[] electors' trust in the Electoral Justice System (+)
\end{itemize} & 
 positively (+) or negatively (-) influence electors' beliefs in the e-voting ecosystem 
 \\ \hline
 
 17 & Mental Moment & 
\begin{itemize}[leftmargin=*]
\item[] positive public perception of IT (+)
\item[] public sentiment of familiarity with IT (+)
\item[] public perception of IT as a major vehicle for prosperity(+)
\item[] public belief that IT is a pre-condition for inclusion in the modern economy (+)
\item[] public perception of trustworthiness of the e-voting system (+)
\item[] public perception of trustworthiness of the electoral authorities (+)
\item[] public perception of e-voting as fraud-free (+)
\item[] electors' belief in e-voting as a democratic process (+)
\item[] public perception of a relaxed atmosphere created by fast and un-crowded voting experience (+)
\item[] public appreciation that the speed of the announcement of the election results increased dramatically with e-voting (+)
\item[] electors' perception about the electoral authorities' ability and willingness to act as an effective guarantor against actions that may compromise the well-functioning of the e-voting ecosystem (+)
\item[] electors' perception of electoral authorities as powerful and benevolent (+)
\item[] Brazilians' constant concern about corruption (-)
\item[] Brazilians’ negative beliefs about the honesty of politicians  (-)
\end{itemize} & 
 positively (+) or negatively (-) influence electors' beliefs about the e-voting ecosystem 
 \\ \hline

18 & Trust Calibration Signals & 
\begin{itemize}[leftmargin=*]
\item[] political parties supporting the e-voting system (+)
\item[] human rights NGOs supporting the e-voting system (in particular, correlating it to transparency and empowerment of the poor) (+)
\item[] media spreading the message of voting as a “right” (+)
\item[] media’s engagement with the effort to inform citizens about where and how to vote (+)
\item[] media advertisements intended to familiarize voters with its technology and procedures (+)
\item[] the use of the voting machine as part of school education and television publicity in the elections period (+)
\item[] Brazilian Government supportive policies, fostering a positive public attitude to IT (+)
\item[] Brazilian Government positive predisposition towards the modernization of elections through IT (+)
\item[] dissemination of the idea of the e-voting system as a mechanism that enabled voting rights, enhancing its image as an institution committed to extending democratic participation (+)
\item[] promotion of fairer elections as the main motivation for introducing e-voting (+)
\item[] approval from parliament and the legal requirement of participation of multiple social actors in the ongoing fine-tuning of the e-voting system and its enactment (+)
\item[] a well-organized and much-publicized process of software enhancements prior to each episode of elections, as well as invitations to political parties and other social actors to participate in test procedures (+)
\item[] publicity of  a clear schedule of activities to test the changes by authorized agencies (+)
\item[] public “ceremony” of the sealing of the e-voting software (+)
\item[] prominent discourse on digital inclusion (+)
\item[] sustained efforts to provide access to IT in poor communities throughout the country, spreading the belief that IT is sine qua non for inclusion in the modern economy (+)
\item[] policies to overcome the digital divide and provide Internet access and IT skills to poor communities in rural and urban areas (+)
\end{itemize} & 
 positively (+) or negatively (-) influence electors' beliefs in the e-voting ecosystem 
 \\ \hline

19 & Trustworthiness Evidence & 
\begin{itemize}[leftmargin=*]
\item[] successful development of the e-voting system and its deployment in efficient and problem-free elections (+)
\item[] e-voting system conception as a democratic instrument, independent of political parties (+)
\item[] electors' successful previous experience with e-voting (+)
\item[] lack of disputes over results (+)
\item[] political parties endorsing the voting arrangements and accepting the election results (+)
\item[] opinion surveys and behavior indicators suggesting that Brazilians in general trust their country’s e-voting process (+)
\item[] positive reputation of the Brazilian IT industry (+)
\item[] activities for upgrading the technology and adjusting the procedures of voting in between elections (+)
\end{itemize} & 
 positively (+) or negatively (-) influence electors' beliefs about the e-voting ecosystem 
 \\ \hline

\end{longtable}

\end{landscape}

\begin{figure*}
\setlength{\fboxsep}{0pt}%
\setlength{\fboxrule}{0pt}%
\subsection*{Appendix 2 - Figures from the ``Trust in artificial intelligence for medical diagnosis'' case}
\end{figure*}

\begin{figure*}
\setlength{\fboxsep}{0pt}%
\setlength{\fboxrule}{0pt}%
\begin{center}
    \centering
    \includegraphics[width=0.9\textwidth]{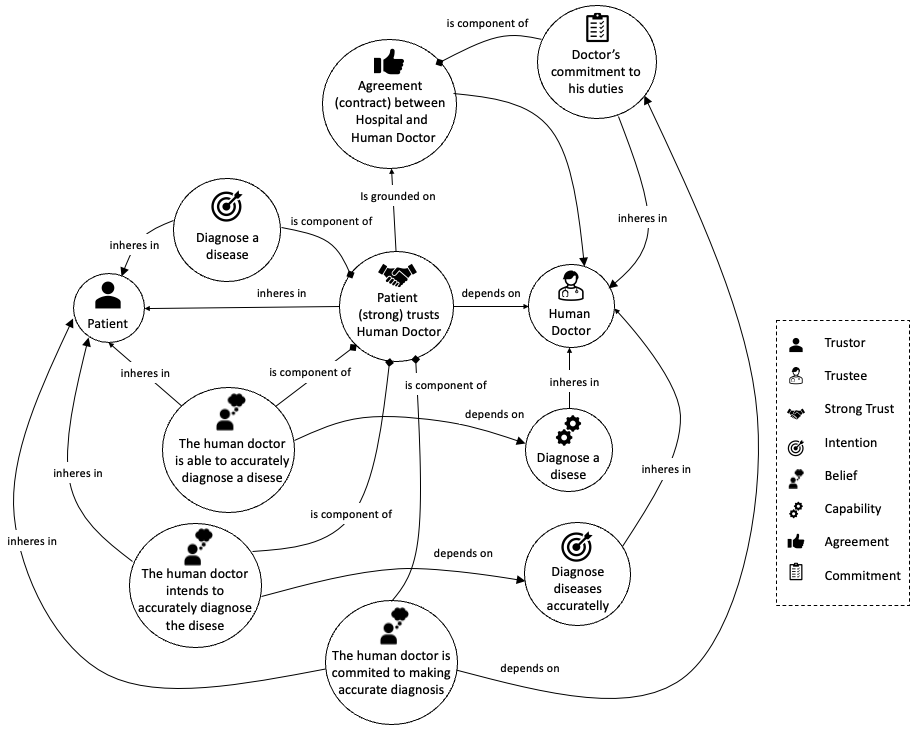}
    \caption{Trust in Human Doctor}
    \label{fig:trust-human}
    \end{center}
\end{figure*}

\begin{figure*}
\setlength{\fboxsep}{0pt}%
\setlength{\fboxrule}{0pt}%
\begin{center}
    \centering
    \includegraphics[width=0.8\textwidth]{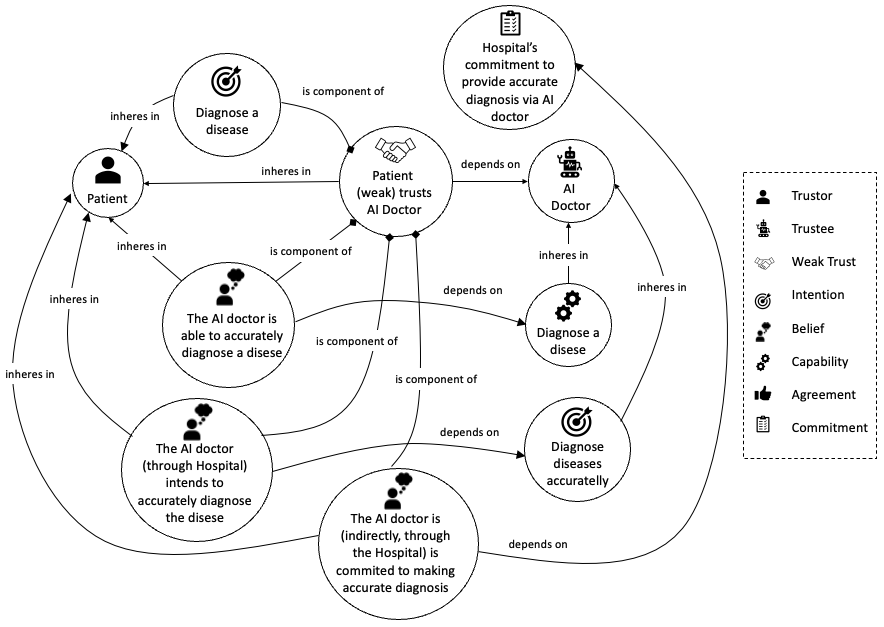}
    \caption{Trust in AI Doctor}
    \label{fig:trust-ai}
    \end{center}
\end{figure*}

\begin{figure*}
\setlength{\fboxsep}{0pt}%
\setlength{\fboxrule}{0pt}%
\begin{center}
    \centering
    \includegraphics[width=0.8\textwidth]{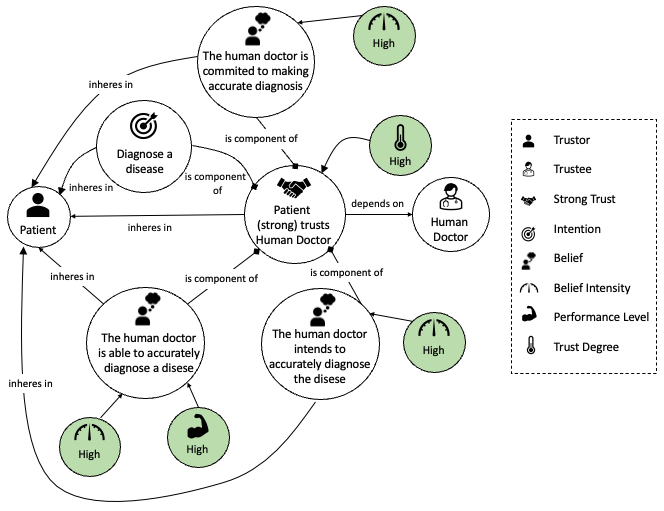}
    \caption{Trust in Human Doctor - Quantitative Perspective}
    \label{fig:quantitative-human}
    \end{center}
\end{figure*}

\begin{figure*}
\setlength{\fboxsep}{0pt}%
\setlength{\fboxrule}{0pt}%
\begin{center}
    \centering
    \includegraphics[width=0.9\textwidth]{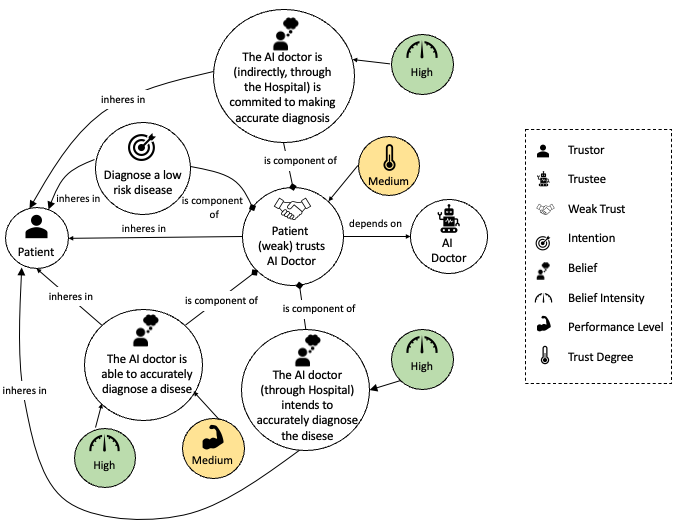}
    \caption{Trust in AI Doctor - Quantitative Perspective}
    \label{fig:quantitative-ai}
    \end{center}
\end{figure*}

\begin{figure*}
\setlength{\fboxsep}{0pt}%
\setlength{\fboxrule}{0pt}%
\begin{center}
    \centering
    \includegraphics[width=0.9\textwidth]{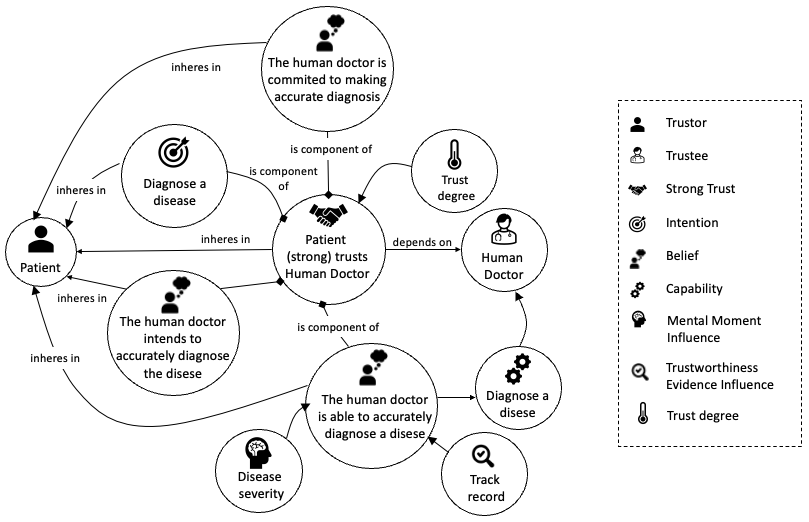}
    \caption{Trust in Human Doctor - Influences}
    \label{fig:influence-human}
    \end{center}
\end{figure*}

\begin{figure*}
\setlength{\fboxsep}{0pt}%
\setlength{\fboxrule}{0pt}%
\begin{center}
    \centering
    \includegraphics[width=0.9\textwidth]{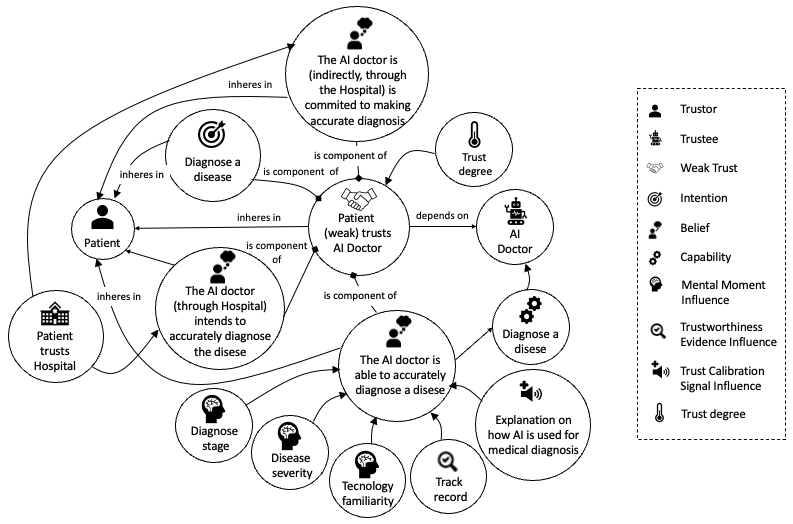}
    \caption{Trust in AI Doctor - Influences}
    \label{fig:influence-ai}
    \end{center}
\end{figure*}

\begin{figure*}
\setlength{\fboxsep}{0pt}%
\setlength{\fboxrule}{0pt}%
\begin{center}
    \centering
    \includegraphics[width=0.9\textwidth]{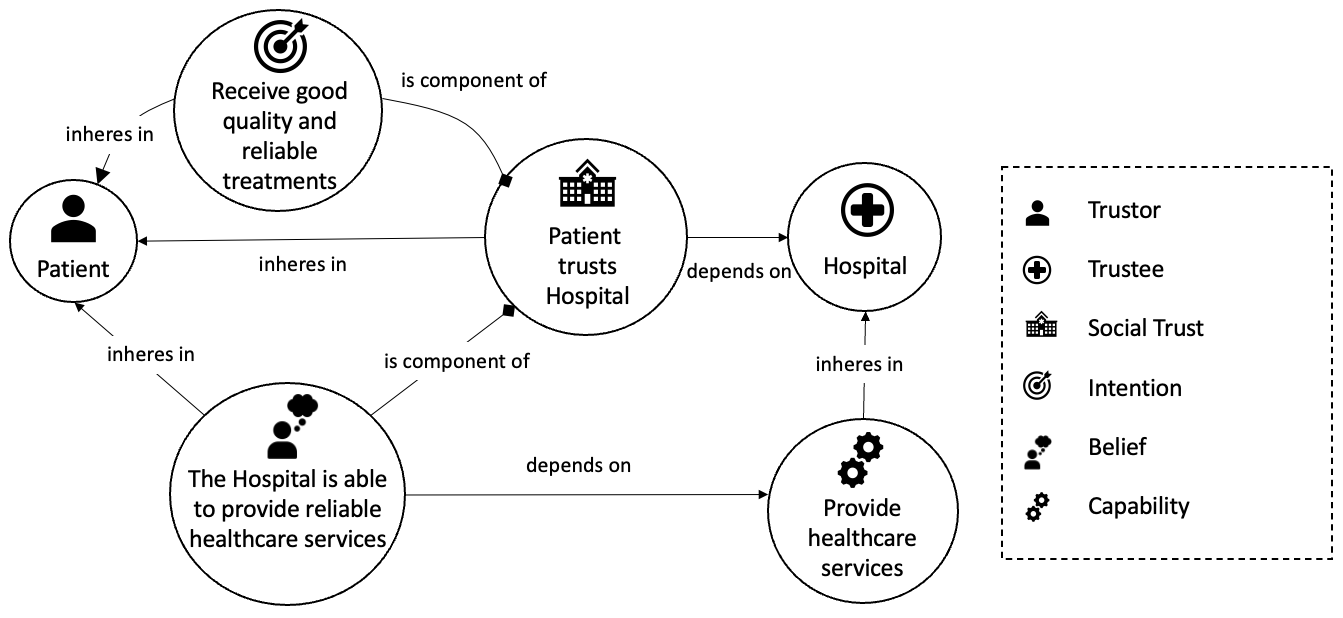}
    \caption{Social Trust in the Hospital}
    \label{fig:trust-institutional}
    \end{center}
\end{figure*}

\end{document}